\documentclass[a4paper, abstract, 11pt, twoside, headinclude]{scrartcl}

\usepackage[T1]{fontenc}
\usepackage[utf8]{inputenc}
\usepackage{lmodern}
\usepackage[english]{babel}
\usepackage{authblk}
\usepackage{stmaryrd}
\usepackage{cite}
\usepackage[headsepline]{scrlayer-scrpage}
\usepackage{standalone}
\usepackage{hyperref}
\usepackage{mathtools}
\usepackage{geometry}
\usepackage{microtype}
\usepackage{nicefrac}
\usepackage{enumitem}
\usepackage[dvipsnames]{xcolor}
\usepackage{xspace}
\usepackage{tikz, pgfplots}
\pgfplotsset{compat=1.15}
\usetikzlibrary{positioning, patterns, calc, angles, quotes, arrows}
\usepackage{amsthm, thmtools, thm-restate}
\PassOptionsToPackage{noabbrev, capitalise, nameinlink}{cleveref}
\input{./abbrv}
\input{./thmdef}
\definecolor{color0}{rgb}{0.12156862745098,0.466666666666667,0.705882352941177} % blue
\definecolor{color1}{rgb}{1,0.498039215686275,0.0549019607843137} % orange
\definecolor{color2}{rgb}{0.172549019607843,0.627450980392157,0.172549019607843} % green
\definecolor{color3}{rgb}{0.83921568627451,0.152941176470588,0.156862745098039} % red
\definecolor{color4}{rgb}{0.580392156862745,0.403921568627451,0.741176470588235} % violet
\definecolor{color5}{rgb}{0.549019607843137,0.337254901960784,0.294117647058824} % brown
\definecolor{color6}{rgb}{0.890196078431372,0.466666666666667,0.76078431372549} % pink
\definecolor{color7}{rgb}{0.737254901960784,0.741176470588235,0.133333333333333} % yellow
\definecolor{color8}{rgb}{0.0901960784313725,0.745098039215686,0.811764705882353} % cyan

\newcommand\marc{}
\def\marc[#1](#2,#3){%
  \pgfmathsetmacro{\angle}{90/#2}
  \draw[#1] (0.0, 0.0) node[fill,circle,minimum size=0.095cm,inner sep=0,outer sep=0] (coord0) {}
  \foreach [count=\k] \x in #3
    {-- ++ (90-\k*\angle:\x) node[fill,circle,minimum size=0.095cm,inner sep=0,outer sep=0] (coord\k) {}}
  ;
}

\newcommand\polychain{}
\def\polychain[#1](#2,#3,#4){
  \edef\angle{90};
  \draw[#1] (#4) node[fill,circle,minimum size=0.075cm,inner sep=0,outer sep=0]{} coordinate (a) ;
  \coordinate (co0) at (a);

  \foreach \x [count=\count,evaluate=\count as \y using {\angle-{#2}[\count-1]}]  in #3 {
    \pgfmathparse{\angle-{#2}[\count-1]};
    \xdef\angle{\pgfmathresult};

    \draw[#1] (a) -- ++(\angle:\x) node[fill,circle,minimum size=0.075cm,inner sep=0,outer sep=0]{} coordinate (a) ;

    \def\this{co\count};
    \coordinate (\this) at (a);
  }
}

\tikzset{%
  add/.style args={#1 and #2}{to path={%
 ($(\tikztostart)!-#1!(\tikztotarget)$)--($(\tikztotarget)!-#2!(\tikztostart)$)%
  \tikztonodes}}
}

\tikzset{
  treenode/.style = {align=center, circle, draw, inner sep=2pt, outer sep=0pt, text centered, minimum width=.75cm},
  mstyle/.style = {treenode},
  vstyle/.style = {circle, draw, fill, minimum width=.05cm, inner sep=.05cm},
}   

\renewenvironment{description}[1][0pt]
  {\list{}{\labelwidth=0pt \leftmargin=#1
   }}
  {\endlist}

\setkomafont{date}{\normalsize}

\setlist[description]{font={\bfseries}}
\setlist[itemize]{label={\raisebox{1pt}{\tiny$\blacksquare$}}}

\makeatletter
\rohead{Jan Macdonald \& Stephan W\"{a}ldchen}
\lehead{\@title}
\hypersetup{
    colorlinks=true,
    linkcolor=NavyBlue,
    filecolor=NavyBlue,
    urlcolor=NavyBlue,
    citecolor=NavyBlue,
    pdftitle={\@title},
    pdfauthor={\@author},
    bookmarks=true,
}
\makeatother
\makeatother

\title{A Complete Characterisation of ReLU-Invariant Distributions}
\author[ ]{Jan Macdonald\footnote{Both authors contributed equally to this work.}}
\author[ ]{Stephan W\"{a}ldchen\protect\footnotemark[1]}
\affil[ ]{Institut f{\"u}r Mathematik, Technische Universit{\"a}t Berlin}
\affil[ ]{\texttt{\{macdonald, stephanw\}@math.tu-berlin.de}}

\begin{document}
 \maketitle

 \begin{abstract}
    We give a complete characterisation of families of probability distributions that are invariant under the action of ReLU neural network layers. The need for such families arises during the training of Bayesian networks or the analysis of trained neural networks, e.g., in the context of uncertainty quantification (UQ) or explainable artificial intelligence (XAI).

    We prove that no invariant parametrised family of distributions can exist unless at least one of the following three restrictions holds: First, the network layers have a width of one, which is unreasonable for practical neural networks. Second, the probability measures in the family have finite support, which basically amounts to sampling distributions. Third, the parametrisation of the family is not locally Lipschitz continuous, which excludes all computationally feasible families.

    Finally, we show that these restrictions are individually necessary. For each of the three cases we can construct an invariant family exploiting exactly one of the restrictions but not the other two.
\end{abstract}

\section{Introduction}

Neural networks have achieved great success in solving diverse problems ranging from image analysis \cite{NIPS2012_4824,NIPS2013_5207}, to natural language processing \cite{cho-etal-2014-learning,NIPS2017_transformer}, to medical applications \cite{annurev-bioeng-071516-044442,MCBEE20181472}.

 An important theoretical as well as computational challenge is calculating the distribution of outputs of a neural network from a given distribution of inputs.
 This task is a crucial ingredient for uncertainty quantification, explainable AI and Bayesian learning.

 More precisely, the task is to describe the distribution $\mu_{\text{out}}$ of
 \begin{equation*}
    f(\bfx)\quad\text{with}\quad\bfx\sim\mu_{\text{in}},
 \end{equation*}
 where $f$ is a neural network function and $\bfx$ is a random input vector distributed according to some probability measure $\mu_{\text{in}}$.
 In other words, we want to determine the push-forward
 \begin{equation}\label{eq:prob_prop}
 \mu_{\text{out}} = f_\ast \mu_{\text{in}}
 \end{equation}
 of $\mu_{\text{in}}$ with respect to the transformation $f$.
 Given the fact that neural networks are highly non-linear functions this generally has no closed form solution.
 Obtaining approximations through numerical integration methods is expensive due to the high dimensionality and complexity of the function.

 This lead to the application of other approximation schemes, such as Assumed Density Filtering \cite{gast-roth-8578453, mwhk-2019-rate-dist} (ADF) and in the case of Bayesian neural networks the more general framework of expectation propagation \cite{jylanki2014expectation, soudry2014expectation} (EP) which contains ADF as a special case. Both methods were originally developed for general large-scale Bayesian inference problems \cite{Minka:2001:FAA:935427, Boyen:1998:TIC:2074094.2074099}.

 Assumed Density Filtering makes use of the feed-forward network structure: If we write $f=f_L\circ f_{L-1}\circ\dots \circ f_2\circ f_1$ as a composition of its layers $f_j$, then \eqref{eq:prob_prop} decomposes as
  \begin{alignat}{3}
    \mu_{0} &= \mu_{\text{in}}, \nonumber\\
    \mu_{j} &= (f_j)_\ast\mu_{j-1} &\quad&\text{for}\quad j=1,\dots,L, \nonumber \\
    \mu_{\text{out}} &= \mu_{L}.\label{eq:layerwise}
 \end{alignat}

 The task can then be described as how to propagate distributions through neural network layers.

 If each $\mu_{j}$ was in the same family of probability distributions as the original $\mu_{\text{in}}$ then we could find efficient layer-wise propagation rules for each layer $f_j$ to obtain $\mu_{\text{out}}$. Such a family, if expressable with a possibly large but finite number of parameters, would be extremely useful for applications involving uncertainty quantification or the need for explainable predictions.

 ADF commonly propagates Gaussians, or more generally exponential families, which are not invariant under the action of neural network layers with non-linear activation. Hence, the method relies on a projection step onto the chosen family of probability distributions after each network layer.
 In other words, \eqref{eq:layerwise} is replaced by
  \begin{alignat*}{3}
    \mu_{0} &= \proj(\mu_{\text{in}}), \\
    \mu_{j} &= \proj((f_j)_\ast\mu_{j-1}) &\quad&\text{for}\quad j=1,\dots,L, \nonumber \\
    \mu_{\text{out}} &= \mu_{L},
 \end{alignat*}
  where $\proj(\cdot)$ is a suitable projection.  In this case only $\mu_{\text{out}}\approx f_{\ast}\mu_{\text{in}}$ holds and there is no guarantee how good the approximation is after multiple layers. This heuristic is justified by the implicit assumption that no invariant families exist.

 Due to the flexibility and richness of the class of neural network functions, it seems intuitively clear that such an invariant family of distributions cannot be realised in a meaningful way---a fact that, to the best of our knowledge, remained unproven. In this paper we will fill this gap.

 \subsection{Contribution}
 We give a complete characterisation of families of probability distributions that are invariant under the action of ReLU neural network layers.
 In fact, we prove that all possible invariant families belong to a set of degenerate cases that are either practically irrelevant or amount to sampling.
 Hence, the implicit assumption made by the ADF and EP frameworks is justified. For each of the degenerate cases we give an explicit example of an invariant family of distributions.

%  We show that there is no need for practitioners to search for ``smarter'' distribution families that can be propagated exactly. Instead, the prime consideration can be numerical convenience.
%  The next step is of course to see if our arguments can be extended to approximately invariant families.

 Our novel proof technique is based on very intuitive geometric constructions and ideas from metric dimension theory. This intuitive argument is made rigorous using properties of the Hausdorff dimension of metric spaces.
 As a result, we provide a theoretical underpinning for statistical inference schemes such as ADF for neural networks.

 \subsection{Applications}

  Calculating output distributions of neural networks is a basic and central problem to make neural networks more applicable. Good performance on test data alone is not sufficient to ensure reliability.
  High-stakes applications such as, e.g., medical imaging or autonomous driving, make questions regarding uncertainty quantification (UQ), robustness, and explainability of network decisions inescapable \cite{pmlr-v136-oala20a, begoli2019need,zhang2018opening,holzinger2017we,michelmore2018evaluating,kim2017interpretable}.

 \paragraph{Uncertainty Quantification}
 Studying the propagation of input uncertainties through an already trained (deterministic) network is used to analyse the reliability of their predictions and their sensitivity to changes in the data. ADF is used towards this goal, e.g., for sigmoid networks for automatic speech recognition \cite{Astudillo2011propagation,Astudillo2014AccountingFT} and ReLU networks for image classification and optical flow \cite{gast-roth-8578453}.

 \paragraph{Explainable Artificial Intelligence}
 So far, neural networks are mostly considered opaque ``black-box'' models. Recent approaches to make them human-interpretable have focused on the crucial subtask of relevance attribution, i.e., finding the most relevant input variables for a given prediction.
 A probabilistic formulation of this task was recently proposed for classification networks \cite{fong2017interpretable, wmhk-2021-jair-compl,mwhk-2019-rate-dist}. Macdonald et al. also considered a heuristic solution strategy for deep ReLU networks that is based on ADF with Gaussian distributions \cite{mwhk-2019-rate-dist}.

  \paragraph{Bayesian Networks}
 Instead of point estimates, Bayesian neural networks (BNN) learn a probability distribution over the network weights through Bayesian inference.
 BNNs provide an inbuilt regularisation by choosing appropriate priors. They naturally lead to network compression when using sparsity-promoting priors or encoding weights of high variance with less precision.
 ADF with Gaussian distributions has been employed for the fast calculation of posteriors during the training of BNNs with sigmoidal \cite{opper1998bayesian} and ReLU \cite{hernandez2015probbackprop} activations.
  A similar approach was taken in \cite{pmlr-v28-wang13a} which relies on layer-wise Gaussian approximations to yield a fast alternative to dropout regularized training.

 \subsection{Outline of the Paper}
 In \cref{sec:prelim_and_result} we state our two main results and provide a high-level description of the proof ideas. In \cref{sec:mainproof} we show that no (non-degenerate) network invariant families of distributions exist. Two constructive steps of the proof are deferred to \cref{sec:xi_surjective,sec:xi_loclip} respectively. Finally, in \cref{sec:counterproof} we prove our second main result by constructing several degenerate invariant families. We conclude with a short discussion in \cref{sec:discussion}.

 \section{Characterisation of Invariant Families of Distributions}\label{sec:prelim_and_result}
 Before we rigorously state and prove our main result, we start by introducing some notation and terminology and precisely specify what we mean by \emph{parametrised families} of probability distributions and by their \emph{invariance} with respect to layers of neural networks.

 \subsection{Preliminaries}
 We denote by $\CB(\R^d)$ the Borel $\sigma$-Algebra on $\R^d$ and by $\CD(\R^d)$ the set of Radon\footnote{In fact $\R^d$ is a separable complete metric space, thus every Borel probability measure is automatically Radon.} Borel probability measures on $\R^d$. We equip $\CD(\R^d)$ with the Prokhorov metric
 \begin{equation*}
     d_P(\mu,\nu) = \inf\{\epsilon >0\,:\, \mu(B) \leq \nu(B^\epsilon) + \epsilon
     \text{ and } \nu(B) \leq \mu(B^\epsilon) + \epsilon \text{ for any }B\in\CB(\R^d)\},
 \end{equation*}
 where $B^\epsilon = \{\,x\in\R^d\,:\, \exists y\in B \text{ with } \|x-y\|<\epsilon\,\}$ is
 the open $\epsilon$-neighbourhood of a Borel set $B\in\CB(\R^n)$.

 \begin{definition}[Family of Distributions]
    Let $\Omega\subseteq\R^n$ and $p\colon\Omega\to\CD(\R^d)$. We call $p$ an $n$-parameter family of probability distributions. It is called a continuous family, if $p$ is continuous.
 \end{definition}
 It might seem more intuitive to refer to the set $\{p(\theta)\}_{\theta\in\Omega}$ as the family of distributions. However, a set of distributions can be parametrised in multiple ways and even the number $n$ of parameters describing such a set is not unique. Whenever we speak about a family of distributions, we never think of it as a mere set of distributions but always attach it to a fixed chosen parametrisation $p$. All our results are stated in terms of this parametrisation.
 \begin{example}[$2$-dimensional Gaussian Family]
 Consider the parameter space
 \[\Omega = \R^2\times\skl{(\sigma_1,\sigma_2,\sigma_3) \in \R^3\,:\,\sigma_1,\sigma_2\geq 0\text{ and }\sigma_1\sigma_2-\sigma_3^2 \geq 0},\]
 and
 \[
  p\colon \Omega \to \CD(\R^2)\colon (\mu_1,\mu_2, \sigma_1, \sigma_2, \sigma_3) \mapsto \CN\kl{\bmat{\mu_1 \\ \mu_2}, \bmat{\sigma_1 & \sigma_3 \\ \sigma_3 & \sigma_2}},
 \]
 where $\CN(\bfmu,\bfSigma)$ denotes the Gaussian distribution with mean $\bfmu$ and covariance matrix $\bfSigma$.
 Then $p$ is a continuous 5-parameter family of 2-dimensional Gaussian distributions.
 \end{example}

 \begin{definition}[Invariance]
    Let $p\colon\Omega\to\CD(\R^d)$ be a family of probability distributions and $f\colon\R^d\to\R^d$ any measurable function. Then the family is called $f$-invariant, if for any $\theta\in\Omega$ the pushforward of the measure $p(\theta)$ under $f$ is again in the family, i.e.\ there exists $\omega\in\Omega$ such that $p(\omega)=f_\ast p(\theta)$. For a collection $\CF\subseteq \{\,f\colon\R^d\to\R^d\,:\,f\text{ measurable}\,\}$ of measurable functions it is called $\CF$-invariant, if it is $f$-invariant for all $f\in\CF$.
 \end{definition}
 For us the special case of invariance with respect to layers of neural networks will be of interest. A neural network is a composition of (possibly many) building blocks, each consisting of an affine transform and a simple componentwise non-linear transform referred to as activation function. These building blocks are also called the layers of a neural network. One frequently used activation function is the rectified linear unit (ReLU) defined as $\varrho(x)=\max\{0,x\}$. Thus, one layer of a ReLU neural network has the general form $\bfx\mapsto\varrho(\bfW\bfx+\bfb)$,
 with so called weight matrix $\bfW$ and bias vector $\bfb$. An $L$-layer ReLU network is the composition of $L$ such functions and has the form
 \[
 \bfx \mapsto \varrho\kl{\bfW_L\varrho\kl{\bfW_{L-1}\varrho\kl{\dots\varrho\kl{\bfW_1\bfx+\bfb_1}\dots} + \bfb_{L-1}} +\bfb_L}.
 \]
 \begin{definition}[ReLU-Invariance]
     Let $p\colon\Omega\to\CD(\R^d)$ be a family of distributions. The family is called ReLU-invariant, if it is $\CF$-invariant for the collection
     \[
     \CF = \{\,f\colon\R^d\to\R^d\colon\bfx\mapsto\varrho(\bfW\bfx+\bfb)\,:\,\bfW\in\R^{d\times d}, \bfb\in\R^d\,\},
     \]
     where $\varrho(x)=\max\{0,x\}$ is applied componentwise.
 \end{definition}
 We observe that if a family $p$ is $f$-invariant and $g$-invariant for two functions $f$ and $g$ such that the composition $g\circ f$ is well-defined, then it is also $(g\circ f)$-invariant. In particular, a ReLU-invariant family is invariant for all ReLU networks of any depths $L$.

\subsection{The Main Result}
 We can now state the main theorem of this work.
 \begin{restatable}{theorem}{mainresult}\label{thm:main}
 Let $\Omega\subseteq\R^n$ and $p\colon\Omega\to\CD(\R^d)$ be a continuous and ReLU-invariant $n$-parameter family of probability distributions. Then at least one of the following restrictions has to hold:
 \begin{description}
     \item[R1. Restricted Dimension:] $d=1$,
     \item[R2. Restricted Support:] $\supp(p(\theta))$ is finite for all $\theta\in\Omega$,
     \item[R3. Restricted Regularity:] $p$ is not locally Lipschitz continuous.
 \end{description}
 \end{restatable}
 This can be interpreted as follows: Besides some rather degenerate cases there can not be any family of probability distributions that is invariant with respect to the layers of a ReLU neural network. The three restrictions characterise which kind of degenerate cases can occur. The first restriction \emph{R1} shows that neural networks with only one neuron per layer are not really powerful function classes, so that the ReLU-invariance is not a strong limitation in this case. However, already two dimensions are enough to unlock the expressive power of neural networks. Restriction \emph{R2} shows that under mild regularity assumptions on the parametrisation the only ReLU-invariant distributions in dimensions $d\geq 2$ are finite mixtures of Dirac distributions. This amounts to sampling distributions of finite size, which of course can work in many scenarios but are often too computationally expensive in high-dimensions. Restriction \emph{R3} shows that continuity alone is not a strong enough assumptions on the parametrisation. This is due to the fact that the class of continuous functions is too flexible and includes non-intuitive examples such as space-filing curves. These can be used to construct examples of invariant distributions. However such a kind of parametrisation has to be be rather wild and would be impractical to work with. A slightly stronger assumption like local Lipschitz continuity is enough to exclude these degenerate examples.

We complement \cref{thm:main} by providing a kind of reverse statement showing that the three restrictions are individually necessary.
 \begin{restatable}{theorem}{maincounter}\label{thm:counter}
    The following statements hold independently of each other.
    \begin{enumerate}[label=(\roman*)]
    \item \label{itm:counter_i} There exists a continuous ReLU-invariant $n$-parameter family of probability distributions that fulfils restriction R1 but not restrictions R2 and R3 from \cref{thm:main}.
    \item \label{itm:counter_ii} There exists a continuous ReLU-invariant $n$-parameter family of probability distributions that fulfils restriction R2 but not restrictions R1 and R3 from \cref{thm:main}.
    \item \label{itm:counter_iii} There exists a continuous ReLU-invariant $n$-parameter family of probability distributions that fulfils restriction R3 but not restrictions R1 and R2 from \cref{thm:main}.
    \end{enumerate}
 \end{restatable}

 \subsection{High-Level Proof Strategy}
  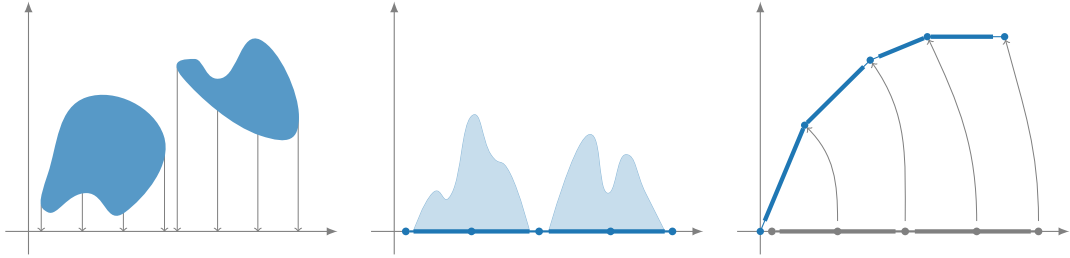
\begin{figure}
    \centering
    \resizebox{\textwidth}{!}{
    \begin{tabular}{ccc}

\begin{tikzpicture}
    \draw[thin, gray, ->, >=latex] (0, -0.3) -- (0, 3);
    \draw[thin, gray, ->, >=latex] (-0.3, 0) -- (4, 0);
    
    \draw[ultra thin, gray, ->, >=to] (0.165, 0.35) -- (0.165, 0);
    \draw[ultra thin, gray, ->, >=to] (0.697, 0.5) -- (0.697, 0);
    \draw[ultra thin, gray, ->, >=to] (1.228, 0.5) -- (1.228, 0);
    \draw[ultra thin, gray, ->, >=to] (1.76, 1) -- (1.76, 0);
    
    \draw[ultra thin, gray, ->, >=to] (1.925, 2.15) -- (1.925, 0);
    \draw[ultra thin, gray, ->, >=to] (2.447, 1.8) -- (2.447, 0);
    \draw[ultra thin, gray, ->, >=to] (2.968, 1.5) -- (2.968, 0);
    \draw[ultra thin, gray, ->, >=to] (3.49, 1.5) -- (3.49, 0);

    \draw[draw=none, fill=color0!75, thick] plot [smooth cycle, tension=0.9] coordinates {(0.25,0.25) (0.75,0.5) (1.25,0.25) (1.75,1.25) (0.75,1.75) (0.25,0.75)} node (A) {};
    
    \draw[draw=none, fill=color0!75, thick] plot [smooth cycle, tension=0.9] coordinates {(2,2) (3,1.25) (3.5,1.5) (3,2.5) (2.5,2) (2.15,2.25)};
    
\end{tikzpicture}
&
\begin{tikzpicture}
    \draw[thin, gray, ->, >=latex] (0, -0.3) -- (0, 3);
    \draw[thin, gray, ->, >=latex] (-0.3, 0) -- (4, 0);
    
    \draw[color0!50] plot [smooth, tension=0.9] coordinates {(0.25,0) (0.5,0.5) (0.75,0.5) (1,1.5) (1.25,1) (1.5,0.75) (1.75,0)};
    \fill[color0!25] plot [smooth, tension=0.9] coordinates {(0.25,0) (0.5,0.5) (0.75,0.5) (1,1.5) (1.25,1) (1.5,0.75) (1.75,0)} -- cycle;
    
    \draw[color0!50] plot [smooth, tension=0.9] coordinates {(2,0) (2.5,1.25) (2.75,0.5) (3,1) (3.25,0.5) (3.5,0)};
    \fill[color0!25] plot [smooth, tension=0.9] coordinates {(2,0) (2.5,1.25) (2.75,0.5) (3,1) (3.25,0.5) (3.5,0)} -- cycle;
    
    \draw[color0, ultra thick] (0.25,0) -- (1.75,0);
    \draw[color0, ultra thick] (2,0) -- (3.5,0);
    
    \draw[color0, thick] (0.2,0) -- (3.6,0);
    \fill[color0] (0.15, 0) circle[radius=0.05];
    \fill[color0] (1, 0) circle[radius=0.05];
    \fill[color0] (1.875, 0) circle[radius=0.05];
    \fill[color0] (2.8, 0) circle[radius=0.05];
    \fill[color0] (3.6, 0) circle[radius=0.05];

\end{tikzpicture}
&
\begin{tikzpicture}
    \draw[thin, gray, ->, >=latex] (0, -0.3) -- (0, 3);
    \draw[thin, gray, ->, >=latex] (-0.3, 0) -- (4, 0);

    \draw[gray, ultra thick] (0.25,0) -- (1.75,0);
    \draw[gray, ultra thick] (2,0) -- (3.5,0);
    
    \draw[gray, thick] (0.2,0) -- (3.6,0);
    \fill[gray] (0.15, 0) circle[radius=0.05] node (precoord0) {};
    \fill[gray] (1, 0) circle[radius=0.05] node (precoord1) {};
    \fill[gray] (1.875, 0) circle[radius=0.05] node (precoord2) {};
    \fill[gray] (2.8, 0) circle[radius=0.05] node (precoord3) {};
    \fill[gray] (3.6, 0) circle[radius=0.05] node (precoord4) {};
    
    \marc[color0](4, {1.5,1.2,0.8,1});
    \draw[color0, ultra thick] ($(coord0)!0.1!(coord1)$) -- (coord1);
    \draw[color0, ultra thick] (coord1) -- ($(coord1)!0.9!(coord2)$);
    \draw[color0, ultra thick] ($(coord2)!0.15!(coord3)$) -- (coord3);
    \draw[color0, ultra thick] (coord3) -- ($(coord3)!0.85!(coord4)$);
    
    \draw[ultra thin, gray, ->, >=to] (precoord1) to[out=90, in=-45] (coord1);
    \draw[ultra thin, gray, ->, >=to] (precoord2) to[out=90, in=-55] (coord2);
    \draw[ultra thin, gray, ->, >=to] (precoord3) to[out=90, in=-65] (coord3);
    \draw[ultra thin, gray, ->, >=to] (precoord4) to[out=90, in=-75] (coord4);
\end{tikzpicture}

\end{tabular}
    }
    \caption{Main steps of transforming a generic probability distribution to a distribution supported on a polygonal chain. The original distribution (left) is projected onto a single dimension and partitioned into intervals (centre) that each contain a sufficiently large portion of the probability mass (shown as the lightly shaded region). Finally it is ``bent'' into a polygonal chain (right). The number of segments of the chain and its segment lengths can be chosen arbitrarily. }
    \label{fig:highlevel}
 \end{figure}

 The main idea for proving \cref{thm:main} is to use the layers of a neural network to transform simple, more or less arbitrary probability distributions to complicated distributions that need an arbitrarily high number of parameters to be described. But a family of parametrised distributions necessarily has a finite number of parameters leading to a contradiction if the family is assumed to be invariant under neural network layers. More precisely, ReLU neural network layers can be used to transform arbitrary probability distributions to distributions that are supported on certain polygonal chains, which we call arcs. This is visualised in \cref{fig:highlevel}. These arcs can be described by the lengths of their line segments. However, the number of segments can be made larger than the number of parameters describing the family of distributions as long as the support of the initial distribution is large enough. From this a contradiction can be derived. There are only three ways to circumvent this, corresponding to the three restrictions in \cref{thm:main}. Firstly, restricting the allowed neural network layers so that transformations to the arcs are not possible. Secondly, restricting the support of the distributions so that the number of line segments they can be transformed to is limited. Thirdly, using a wild parametrisation that leverages ideas similar to space-filling curves in order to use only few parameters to describe a set that effectively would require more parameters.

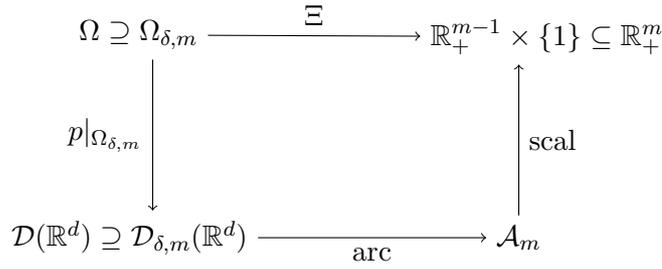
\begin{figure}
    \centering
    % \scriptsize

\begin{tikzpicture}

\node (Omega) {$\Omega\supseteq\Omega_{\delta, m}$};
\node[below=2cm of Omega.300, anchor=50] (D) {$\CD(\R^d)\supseteq\CD_{\delta,m}(\R^d)$};
\node[right=3cm of D] (A) {$\CA_m$};
\node[above=2cm of A, anchor=225] (Rm) {$\R_+^{m-1}\times\{1\}\subseteq\R_+^m$};

\draw[->] (Omega.300) -- (D.50) node[midway, left] {$p\vert_{\Omega_{\delta,m}}$};
\draw[->] (D) -- (A) node[midway,below] {$\arc$};
\draw[->] (A) -- (Rm.225) node[midway, right] {$\scale$};
\draw[->] (Omega) -- (Rm) node[midway, above] {$\Xi$};

\end{tikzpicture}
    \caption{Schematic overview of the spaces and functions involved in our proof. }
    \label{fig:overview}
 \end{figure}

 The idea of mapping (a subset of) the parametrisation domain to distributions supported on polygonal arcs, then to the respective arcs, and finally to segment lengths is schematically shown in \cref{fig:overview}. We will give exact definitions of all involved spaces and mappings shortly.
 The full proof for \cref{thm:main} will be given in \cref{sec:mainproof} and the proof for the reverse \cref{thm:counter} in \cref{sec:counterproof}.

\section{Proof of the Main Result (\cref{thm:main})}\label{sec:mainproof}
We will now present the main steps of the proof of \cref{thm:main}. The proofs of the individual steps will be deferred to later sections in order to not distract from the main idea. We will assume towards a contradiction that all three restrictions in \cref{thm:main} are not satisfied.
\begin{assumption}\label{asm:main}
Let $d\geq 2$, $\Omega\subseteq\R^n$, and $p\colon\Omega\to\CD(\R^d)$ be a locally Lipschitz continuous and ReLU-invariant $n$-parameter family of probability distributions such that there exists a $\theta\in\Omega$ for which $\supp\kl{p(\theta)}$ is not finite.
\end{assumption}
From this we will derive a contradiction. The proof is based on the idea of transforming arbitrary probability distributions to distributions supported on certain polygonal chains, which we call arcs.
\begin{definition}[Arcs]\label{def:arc}
    An $m$-arc is a subset $A\subseteq \mathbb{R}^2$ defined as a polygonal chain, i.e. a connected piecewise linear curve,
     \[
        A = \bigcup_{i=1}^m \conv\kl{\skl{\bfv_{i-1}, \bfv_i}},
    \]
    determined by $m+1$ vertices $\{\bfv_0, \dots, \bfv_m\}$, that satisfy
    \[
     \bfv_i = \bfv_{i-1} + r_i \begin{bmatrix}\sin(i \phi_m)\\ \cos(i \phi_m)\end{bmatrix}, \quad i=1,\dots,m
    \]
    for some length scales $r_1,\dots, r_m > 0$ and the angle $\phi_m = \frac{\pi}{2m}$.
\end{definition}
We denote the vertex set of an $m$-arc $A$ as $\vertex(A)=\{\bfv_0,\dots,\bfv_m\}$, the set of its line segments as $\segment(A) = \{\ell_1,\dots,\ell_m\}$, where $\ell_i=\conv\kl{\skl{\bfv_{i-1}, \bfv_i}}$, and the set of its scaling factors as $\scale(A)=\{r_1,\dots,r_m\}$.

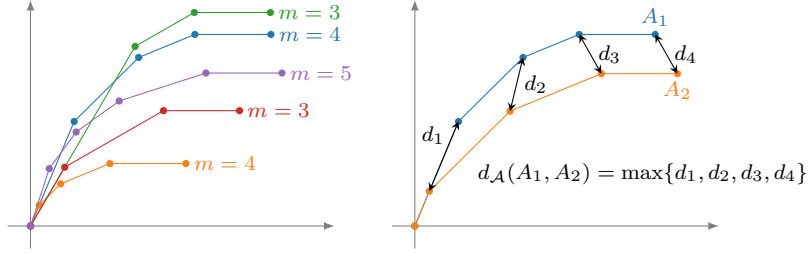
\begin{figure}
     \centering
     \scriptsize

\begin{tabular}{cc}
\begin{tikzpicture}
    \draw[thin, gray, ->, >=latex] (0, -0.3) -- (0, 3);
    \draw[thin, gray, ->, >=latex] (-0.3, 0) -- (4, 0);
     \marc[color0](4, {1.5,1.2,0.8,1.0});
     \node[right,color0] at (coord4) {$m=4$};
     \marc[color1](4, {0.3,0.4,0.7,1.0});
     \node[right,color1] at (coord4) {$m=4$};
     \marc[color2](3, {2.75,0.9,1.0});
     \node[right,color2] at (coord3) {$m=3$};
     \marc[color3](3, {0.9,1.5,1.0});
     \node[right,color3] at (coord3) {$m=3$};
     \marc[color4](5, {0.8,0.6,0.7,1.2,1.0});
     \node[right,color4] at (coord5) {$m=5$};
    
\end{tikzpicture}
&
\begin{tikzpicture}
    \draw[thin, gray, ->, >=latex] (0, -0.3) -- (0, 3);
    \draw[thin, gray, ->, >=latex] (-0.3, 0) -- (4, 0);
    
    \marc[color0](4, {1.5,1.2,0.8,1.0});
    \coordinate (A11) at (coord1);
    \coordinate (A12) at (coord2);
    \coordinate (A13) at (coord3);
    \coordinate (A14) at (coord4);
    \node[above,color0] at (A14) {$A_1$};
    
    \marc[color1](4, {0.5,1.5,1.3,1.0});
    \coordinate (A21) at (coord1);
    \coordinate (A22) at (coord2);
    \coordinate (A23) at (coord3);
    \coordinate (A24) at (coord4);
    \node[below,color1] at (A24) {$A_2$};
    
    \draw[<->, >=stealth] (A11) -- (A21) node[pos=.2,left] {$d_1$};
    \draw[<->, >=stealth] (A12) -- (A22)  node[midway,right] {$d_2$};
    \draw[<->, >=stealth] (A13) -- (A23)  node[midway,right] {$d_3$};
    \draw[<->, >=stealth] (A14) -- (A24)  node[midway,right] {$d_4$};
    
    \node at (3,0.7) {$d_\CA(A_1,A_2)=\max\{d_1,d_2,d_3,d_4\}$};
\end{tikzpicture}
\end{tabular}
     \caption{Examples of several standard $m$-arcs for varying $m$ (left). Standardised arcs are always contained within the non-negative orthant and end on a horizontal line segment. The distance between two $m$-arcs is the maximal Euclidean distance of corresponding vertices (right).}
     \label{fig:arc_examples}
 \end{figure}

It will turn out to be useful to remove some ambiguity from the set of $m$-arcs by standardising their behaviour at the start and end vertices. Some examples can be seen in \cref{fig:arc_examples}.
\begin{definition}[Standard Arcs]
 An $m$-arc $A$ with vertices $\skl{\bfv_0,\dots,\bfv_m}$ and length scales $\skl{r_1,\dots,r_m}$ is called standardised (or a standard $m$-arc), if it starts at the origin and has a normalised last line segment, i.e.\ if $\bfv_0=\bfzero$ and $r_m=1$. The set of all standard $m$-arcs is denoted $\CA_m$ and equipped with the metric
 \[
    d_{\CA}(A_1, A_2) = \max_{i=1,\dots,m} \|\vertex\kl{A_1}_i - \vertex\kl{A_2}_i\|_2.
 \]
\end{definition}

 The metric $d_\CA$ is induced by a mixed $(\ell_2,\ell_\infty)$-norm and thus indeed a proper metric. There is a one-to-one correspondence between the sets $\CA_m$ and $\R^{m-1}_+\times\skl{1}\cong\R^{m-1}_+$ via the scaling factors.

 \begin{definition}[Arc-Supported Measures]
 A probability measure $\mu\in\CD(\R^2)$ is said to be $\delta$-distributed on a standard $m$-arc $A\in\CA_m$ if $\supp(\mu) \subseteq A$ and for each line segment $\ell\in\segment(A)$ we have $\mu(\ell) \geq \delta$.

 For $d>2$, a probability measure $\mu\in\CD(\R^d)$ is said to be $\delta$-distributed on a standard $m$-arc $A\in\CA_m$ if $\supp(\mu) \subseteq \linspan\{\bfe_1,\bfe_2\}$ and by identifying $\linspan\{\bfe_1,\bfe_2\}$ with $\R^2$ it is $\delta$-distributed on a standard $m$-arc in the above sense.

 We denote the set of probability measures that are $\delta$-distributed on standard $m$-arcs by $\CD_{\delta,m}(\R^d)\subseteq\CD(\R^d)$.
 \end{definition}

\begin{lemma}\label{lem:arc_unique}
    Let $\mu\in\CD_{\delta,m}(\R^d)$ be $\delta$-distributed on a standard $m$-arc, then this arc is unique.
\end{lemma}
\begin{proof}
 Towards a contradiction assume that $\mu$ is $\delta$-distributed on $A_1, A_2 \in \CA_m$ and $A_1 \neq A_2$. Let $\skl{\ell_1^j,\dots,\ell_m^j} = \segment(A_j)$ and $\skl{r_1^j,\dots,r_m^j} = \scale(A_j)$ denote the line segments and scaling factors of $A_1$ and $A_2$ respectively.
 Since $A_1\neq A_2$ and $r^1_m=r^2_m=1$ there is a smallest index $1\leq i\leq m-1$ such that $r_i^1\neq r^2_i$. Without loss of generality assume $r_i^1 < r^2_i$. Then $\ell^2_{i+1}$ lies outside of the convex hull of $A_1$ and in particular
 \[
 \ell^2_{i+1} \cap A_1 = \emptyset.
 \]
 But this contradicts the fact that both $\mu(\ell^2_{i+1})\geq \delta$ and $\supp\kl{\mu}\subseteq A_1$ must hold.
 \end{proof}

\Cref{lem:arc_unique} shows that arc-supported measures induce unique standard $m$-arcs. We denote the corresponding function mapping a measure $\mu\in\CD_{\delta,m}(\R^d)$ to its induced arc $A_\mu\in\CA_m$ by $\arc\colon\CD_{\delta,m}(\R^d)\to\CA_m\colon\mu\mapsto A_\mu$.
For an $n$-parameter family of distributions, $p\colon\Omega\to\CD(\R^d)$, we denote by
\[
    \Omega_{\delta, m}=\skl{\,\theta\in\Omega\,:\,p(\theta)\in\CD_{\delta,m}(\R^d)\,}\subseteq\Omega
\]
the set of parameters mapping to arc-supported measures. Without further assumptions on $p$ this set might well be empty. We will see however, that this can not happen for ReLU-invariant families as long as they contain at least one measure with a support of cardinality at least $m$.

We are now ready to start discussing the main steps of the proof of \cref{thm:main}. It relies on the following three results.

\begin{restatable}[Surjectivity]{lemma}{Surjectivity}\label{lem:xi_surjective}
Under \cref{asm:main} and for any $m\in\N$ there exists a $\delta>0$ such that the map $\Xi\colon\Omega_{\delta,m}\subseteq\R^n\to\R^{m-1}_+\times\{1\}$ given by $\Xi=\scale\circ\arc\circ\,p\vert_{\Omega_{\delta,m}}$ is surjective. In particular, in this case the domain $\Omega_{\delta,m}$ is non-empty.
\end{restatable}
The proof for this is given in \cref{sec:xi_surjective}. In short, we show that any measure with non finite support can be transformed by ReLU layers into a measure supported on an arbitrary standard arc. Then using the scaling factors to identify arcs with $\R^{m-1}_+\times\{1\}$ yields the claim.

\begin{restatable}[Local Lipschitz continuity]{lemma}{LocLip}\label{lem:xi_loclip}
Under \cref{asm:main} and for any $m\in\N$ let $\delta>0$ and $\Xi\colon\Omega_{\delta,m}\subseteq\R^n\to\R^{m-1}_+\times\{1\}$ be as in \cref{lem:xi_surjective}. Then $\Xi$ is locally Lipschitz continuous.
\end{restatable}
The proof for this is given in \cref{sec:xi_loclip}. In short, we show that all three partial functions $\scale$, $\arc$, and $p$ are locally Lipschitz continuous, hence also their composition $\Xi$ is.

As a final piece before the main theorem we need a rather general result on maps between Euclidean spaces.
\begin{restatable}{lemma}{hdimlemma}\label{lem:loclip_n_to_m}
Let $B\subseteq \R^n$ and $f\colon B\to \R^m$ be locally Lipschitz continuous. If the image $f(B)$ is a Borel set with non-empty interior in $\R^m$, then $m\leq n$. In particular, if $f$ maps surjectively onto $\R^m$ or $\R^m_+$, then $m\leq n$.
\end{restatable}
The full proof of this is given in \ref{apx:hdim}. It essentially uses the fact that locally Lipschitz continuous maps cannot increase the Hausdorff dimension. Since the Hausdorff dimension of $B$ is at most $n$ and the Hausdorff dimension of $f(B)$ is $m$, the claim follows.

We now have all the ingredients to prove the main theorem.
\begin{proof}[Proof of \cref{thm:main}]
Towards a contradiction let \cref{asm:main} hold. Let $m>n+1$ be arbitrary. By \cref{lem:xi_surjective} there exists a $\delta>0$ so that $\Omega_{\delta,m}$ is non-empty and  $\Xi\colon\Omega_{\delta,m}\to\R^{m-1}_+\times\{1\}$ is surjective. By \cref{lem:xi_loclip} it is also locally Lipschitz continuous. Since $\R^{m-1}_+\times\{1\}$ is isometric to $\R^{m-1}_+$ and $m-1>n$ this contradicts \cref{lem:loclip_n_to_m}. Hence, one of the assumptions in \cref{asm:main} cannot hold, which proves \cref{thm:main}.
\end{proof}

 \section{Proof of \cref{lem:xi_surjective}: Surjectivity of \texorpdfstring{$\Xi$}{Xi}}\label{sec:xi_surjective}

 For the remainder of the section let \cref{asm:main} hold. We will prove \cref{lem:xi_surjective} and show that for any $m\in\N$ there exists a $\delta>0$ so that the map $\Xi\colon\Omega_{\delta,m}\to\R^{m-1}_+\times\{1\}$ is surjective. We will do this by transforming (through a series of ReLU layers) any measure with infinite support into a measure that is supported on an arbitrary standard $m$-arc. Such an infinitely supported measure exists within the $n$-parameter family $p$ by assumption. Due to the ReLU-invariance the transformed measure is then also included in the family. As a consequence $\Omega_{\delta,m}$ must be non-empty and in fact we show that $\arc\circ\,p\vert_{\Omega_{\delta,m}}\colon\Omega_{\delta,m}\to\CA_m$ is surjective. The choice of $\delta$ will depend on how ``evenly spread'' the support of the original, untransformed, measure is. Finally, we have already discussed that $\CA_m$ can be identified with $\R^{m-1}_+\times\{1\}$ via the scaling factors.

 As a first step we observe how transforming measures by continuous functions affects their support. The support $\supp(\mu)$ of a Radon measure $\mu$ is defined as the complement of the largest open $\mu$-null set. A point $\bfx$ belongs to $\supp(\mu)$ if and only if every open neighbourhood of $\bfx$ has positive measure. A Borel set contained in the complement of $\supp(\mu)$ is a $\mu$-null set. The converse holds for open sets, i.e.\ an open set intersecting $\supp(\mu)$ has positive measure (in fact it suffices if the intersection is relatively open in $\supp(\mu)$).
 \begin{lemma}\label{lem:cont_supp}
 For any $q,r\in\N$ let $\mu\in\CD(\R^q)$ and let $f\colon\R^q\to\R^r$ be continuous. Then
 \[
    \supp\kl{f_\ast(\mu)} = \overline{f\kl{\supp(\mu)}},
 \]
 where $f_\ast(\mu)\in\CD(\R^r)$ is the pushforward of $\mu$ under $f$.
 \end{lemma}
 \begin{proof}
 We begin by showing $f\kl{\supp(\mu)}\subseteq\supp\kl{f_\ast(\mu)}$. Let $\bft\in f\kl{\supp(\mu)}$ and $U\subseteq\R^r$ be any open neighbourhood of $\bft$. There exists $\bfx\in\supp(\mu)$ so that $\bft=f(\bfx)$. Since $\bft\in U$, we get $\bfx\in f^{-1}(U)$. Thus $f^{-1}(U)$ is an open neighbourhood of $\bfx$ and since $\bfx\in\supp(f)$ we get $f_\ast(\mu)(U) = \mu(f^{-1}(U))>0$. Since $U$ was an arbitrary neighbourhood of $\bft$ this shows $\bft\in\supp\kl{f_\ast(\mu)}$. This implies $f\kl{\supp(\mu)}\subseteq\supp\kl{f_\ast(\mu)}$. But $\supp\kl{f_\ast(\mu)}$ is closed, so we can even conclude $\overline{f\kl{\supp(\mu)}}\subseteq\supp\kl{f_\ast(\mu)}$. \par
 We will show the converse inclusion by showing $\overline{f\kl{\supp(\mu)}}^c\subseteq\supp\kl{f_\ast(\mu)}^c$. For this let $\bft\in \overline{f\kl{\supp(\mu)}}^c$. Then there exists an open neighbourhood $U$ of $\bft$ such that $U\cap \overline{f\kl{\supp(\mu)}} = \emptyset$, which implies $f^{-1}(U)\cap\supp(\mu)=\emptyset$. But this is only possible if $f_\ast(\mu)(U)=\mu\kl{f^{-1}(U)}=0$ and therefore $\bft\in\supp\kl{f_\ast(\mu)}^c$.
 \end{proof}

 To simplify notations in the following, we begin the transformation by establishing a standardised situation that takes places exclusively in the two-dimensional subspace $\linspan\skl{\bfe_1,\bfe_2}$ of $\R^d$. In fact, all $m$-arcs are essentially one-dimensional objects embedded in a two-dimensional space. Therefore, we start by transforming the infinitely supported measure to the non-negative part of the one-dimensional space $\linspan\skl{\bfe_1}$, which will then subsequently be ``bent'' into the correct arc living in $\linspan\skl{\bfe_1,\bfe_2}$.

\subsection{Projection to a  One-Dimensional Subspace}\label{sec:onedim_twodim}

 \begin{lemma}\label{lem:proj_one_dim}
 Let $\mu\in\CD(\R^d)$ such that $\supp(\mu)$ is not finite. Then there exists $\bfW\in\R^{d\times d}$ and $\bfb\in\R^d$ such that the image of $f\colon\R^d\to\R^d\colon\bfx\mapsto\varrho(\bfW\bfx+\bfb)$ is contained in $\linspan\skl{\bfe_1}$ and $\supp\kl{f_\ast (\mu)}$ is infinite.
 \end{lemma}
 \begin{proof}
  Consider the coordinate projections $\proj_i\colon\R^d\to\R\colon\bfx\mapsto\bfe_i^\top\bfx$. Clearly we have
  \[
   \supp(\mu) \subseteq \proj_1\kl{\supp(\mu)}\times\dots\times\proj_d\kl{\supp(\mu)}.
  \]
  By assumption $\supp(\mu)$ is infinite and the product can only be infinite if at least of its factors is infinite, for example $\proj_j\kl{\supp(\mu)}$. If $\proj_j\kl{\supp(\mu)}\cap\R_+$ is infinite, we set $\sigma=+1$, otherwise $\proj_j\kl{\supp(\mu)}\cap\R_-$ must be infinite and we set $\sigma=-1$. Now choosing
  \[
    \bfW = \begin{bmatrix} \sigma\bfe_j & \bfzero & \dots & \bfzero\end{bmatrix}^\top\quad\text{and}\quad
    \bfb = \bfzero,
  \]
  clearly yields $f(\R^d)\subseteq\linspan\skl{\bfe_1}$. Further, by \cref{lem:cont_supp} we have \begin{align*}
    \supp\kl{f_\ast(\mu)} &= \overline{f\kl{\supp(\mu)}} \\
    &= \overline{\varrho\kl{\sigma \proj_j\kl{\supp(\mu)}}}\times\skl{0}\times\dots\times\skl{0} \\
    &= \begin{cases}
        \overline{\proj_j\kl{\supp(\mu)}\cap\R_+}\times\skl{0}\times\dots\times\skl{0}, &\quad\text{if }\sigma=+1\\
        \overline{-\proj_j\kl{\supp(\mu)}\cap\R_+}\times\skl{0}\times\dots\times\skl{0}, &\quad\text{if }\sigma=-1
       \end{cases},
  \end{align*}
  which by the choice of $\sigma$ is infinite.
 \end{proof}
 From now on we only need to consider measures with infinite support within the non-negative part of $\linspan\{\bfe_1\}$. Two dimensions will be enough for our construction to ``bend'' the measures into arc shape. Any two-dimensional ReLU layer $\R^2\to\R^2\colon\bfx\mapsto\varrho(\bfW\bfx+\bfb)$ can easily be extended to a $d$-dimensional ReLU layer
 \[
    \R^d\to\R^d\colon\bfx\mapsto\varrho\kl{\begin{bmatrix} \bfW & \bfzero_{2\times(d-2)} \\ \bfzero_{(d-2)\times 2} & \bfzero_{(d-2)\times(d-2)} \end{bmatrix}\bfx + \begin{bmatrix} \bfb \\ \bfzero_{(d-2)\times 1}\end{bmatrix}},
 \]
 only operating on $\linspan\skl{\bfe_1,\bfe_2}\subseteq\R^d$ by appropriately padding with zeros. To simplify the notation, we leave out these zero paddings and just assume without loss of generality $d=2$ from now on.

 \subsection{Partitioning into Line Segments}

Now we show how the infinite support on the non-negative real line can be partitioned into an arbitrary number of intervals, which in the end will correspond to the arc line segments. We can write $\supp(\mu) = S\times\skl{0}\subseteq\R^2$ for some set $S\subseteq\R_{\geq 0}$.
 \begin{lemma}\label{lem:interval_partition}
 Let $S\subseteq\R_{\geq0}$ be infinite. Then for any $m\in\N$ there exist distinct points $0=b_0<\dots<b_m<\infty$ such that $S\cap \kl {b_{j-1},b_{j}}\neq\emptyset$ for all $j=1,\dots,m$.
 \end{lemma}
 \begin{proof}
 Since $S$ is infinite for any $m\in\N$ there exist distinct points $s_1,\dots,s_m\in S$ with $0<s_1<\dots<s_m$. Setting $b_0=0$,
 \[
    b_j = \frac{s_j+s_{j+1}}{2}\quad\text{for}\quad j=1,\dots,m-1,
 \]
 and $b_m = s_m + 1$ we get $s_j\in\kl{b_{j-1},b_j}$ for $j=1,\dots,m$.
 \end{proof}
 Further, we can restrict the support of the measure to a compact subset by clipping  the set $S$ to the range $\ekl{b_0,b_m}=\ekl{0,b_m}$. This can be achieved using the two layer ReLU network
 \[
 \R^2\rightarrow\R^2 \colon \bfx \mapsto \varrho\kl{-\varrho\kl{-\bfx+ \begin{bmatrix} b_m \\ 0 \end{bmatrix}} + \begin{bmatrix} b_m \\ 0\end{bmatrix}}
 \]
 with $b_m$ as in \cref{lem:interval_partition}. Altogether, without loss of generality we can from now on assume that $\mu$ is supported on $[b_0,b_m]\times\skl{0}\subseteq\R^2$ and the support intersects each subset $\kl{b_{j-1},b_j}\times\skl{0}$ for $j=1,\dots,m$.

 \subsection{Resizing the Line Segments}

 After partitioning the support of a distribution into intervals that each contain a certain amount of the probability mass we now want to bend it into arc shape. Each interval in the partition will correspond to a line segment. However, the bending will distort the segment lengths, so in order to obtain an arc with specified segment lengths we first have to resize the intervals. This is also possible with ReLU layers.

  \begin{lemma}\label{lem:nn_relu_b2a}
 For $m\geq 2$ let $0=a_0<\dots<a_m<\infty$ and $0=b_0<\dots<b_m<\infty$. Then there exists a collection of ReLU layer transformations $\skl{F_j\colon\R^2\to\R^2}_{j=0,...,m}$ such that their composition $F = F_{m}\circ F_{m-1}\circ \dots \circ F_{1}\circ F_{0}$ satisfies
 \[
    F\kl{\bmat{b_i \\ 0}} = \bmat{a_i \\ 0}\quad\text{for all}\quad  i=0,\dots,m,
\]
and restricted to $\R_{\geq 0}\times\{0\}$ it is a piecewise linear map with breakpoints (possibly) at $b_0,\dots,b_m$.
 \end{lemma}

\begin{proof}

 The proof consists of two parts. First, we iteratively construct a collection of piecewise linear and strictly increasing functions $\skl{f_j\colon\R\to\R}_{j=0,...,m}$ such that
 \begin{equation}\label{eq:resize_segments}
    \kl{f_j\circ\dots\circ f_0}\kl{b_i} = a_i \quad\text{for all}\quad i=0,\dots,j,
 \end{equation}
 for any $j=0,\dots,m$. In particular for $j=m$ and $f=f_m\circ f_{m-1}\circ\dots\circ f_0$ we obtain $f(b_i)=a_i$ for all $i=0,\dots,m$.
 Afterwards we show how to construct $\skl{F_j}_j$ from $\skl{f_j}_j$.

 Since $b_0=a_0=0$, we can start with
 \[
   f_0(x) = x
 \]
 and see that \eqref{eq:resize_segments} is satisfied for $j=0$. Now assuming we have already constructed strictly increasing and piecewise linear functions $f_0,\dots,f_j$ satisfying \eqref{eq:resize_segments} we now want to construct $f_{j+1}$. The idea is to choose it in such a way that it leaves the points already correctly mapped by $f_j\circ\dots\circ f_0$ unchanged but linearly transforms the remaining points so that $b_{j+1}$ is mapped to $a_{j+1}$. For brevity we denote $b_i^j = \kl{f_j\circ\dots\circ f_0}(b_i)$ for $i=0,\dots,m$. By assumption $b_i^j=a_i$ for $i=0,\dots,j$ and by strict monotonicity $b_{j+1}^j>b_j^j=a_j$. We set
 \[
   f_{j+1}(x)
   = \begin{cases}
   x, &\quad x \leq a_j, \\
   %a_j + \frac{a_{j+1}-a_j}{b^j_{j+1}-a_j}(x-a_j) =
   x + \frac{a_{j+1}-b^j_{j+1}}{b^j_{j+1}-a_j}(x-a_j), &\quad x > a_j.
   \end{cases}
 \]
 It is easy to see that this function leaves all points $a_1<a_2\dots<a_j$ unchanged, maps $b^{j}_{j+1}$ to $a_{j+1}$ and is continuous, monotonously increasing and piecewise linear.
 We can also express $f_{j+1}$ using $\varrho$, i.e.
 \begin{equation}\label{eq:fj_resize}
    f_{j+1}(x) = x + \frac{a_{j+1} - b^j_{j+1}}{b^j_{j+1} - a_j} \varrho\kl{x-a_j}.
 \end{equation}
 Then $f_{j+1}$ maps $b^j_{j+1}$ to $a_{j+1}$ and acts as the identity map on the region $x\leq a_j$ where points are already correctly mapped, as illustrated in \ref{fig:mapping}.  Hence clearly \eqref{eq:resize_segments} is also satisfied for $j+1$. To see that $f_{j+1}$ is strictly increasing we observe that its slope is $1$ on the region $x\leq a_j$ and
 \[
    1+\frac{a_{j+1} - b^j_{j+1}}{b^j_{j+1} - a_j} = \frac{a_{j+1} - a_j}{b^j_{j+1} - a_j} >0
 \]
 on the region $x\geq a_j$. This adds at most one new breakpoint at $b_j$ to the composition $f_{j+1}\circ\dots\circ f_0$. Continuing this process until $j=m$ finishes the first part.
 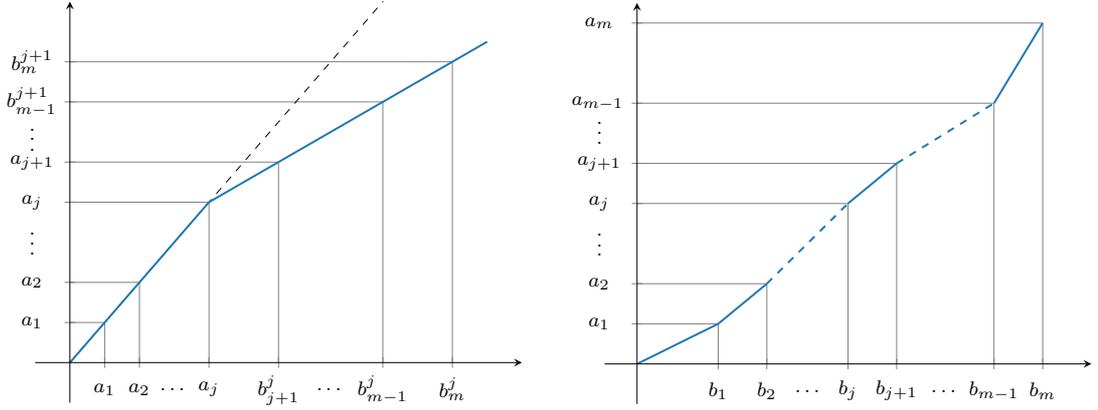
\begin{figure}
     \centering
     \resizebox{\textwidth}{!}{
     \scriptsize

% func(x, a, w)
\pgfmathdeclarefunction{func}{3}{%
\pgfmathparse{%
(#1<=#2) * (#1) +%
(#1>#2) * (#2+#3*(#1-#2))%
}%
}

\begin{tabular}{@{}c@{\qquad}c@{}}
\begin{tikzpicture}
    \begin{axis}[
        axis on top=true,
        axis x line=middle,
        axis y line=middle,
        ymin=-0.5, ymax=4.5,
        ytick=data,
        yticklabels={$a_1$, $a_2$, $a_j$, $a_{j+1}$, $b_{m-1}^{j+1}$, $b_m^{j+1}$},
        xmin=-0.5, xmax=6.5,
        xtick=data,
        tick style={line width=0.25, gray},
        tick label style={minimum width=3em, minimum height=2em},
        xticklabels={$a_1$, $a_2$, $a_j$, $b^j_{j+1}$, $b_{m-1}^j$, $b_m^j$},
        extra x ticks={1.5, 3.75},
        extra x tick labels={$\dots$, $\dots$},
        extra x tick style={tick style={draw=none}},
        extra y ticks={1.575, 2.875},
        extra y tick labels={$\vdots$, $\vdots$},
        extra y tick style={tick style={draw=none}},
    ]
    \addplot[draw=none, samples at={0.5, 1.0, 2.0, 3.0, 4.5, 5.5}]{func(x,2,0.5)};

    \pgfplotsinvokeforeach{0.5, 1.0, 2.0, 3.0, 4.5, 5.5}{
        \draw[line width=0.25, gray] (#1,{func(#1,2,0.5)}) -- (#1,0);
        \draw[line width=0.25, gray] (#1,{func(#1,2,0.5)}) -- (0,{func(#1,2,0.5)});
    }

    \addplot[line width=0.4, dashed, samples at={0.0,6}]{\x};
    \addplot[thick, color0, samples at={0.0, 0.5, 1.0, 2.0, 3.0, 4.5, 5.5, 6}]{func(x,2,0.5)};
    \end{axis}

\end{tikzpicture}
&
\begin{tikzpicture}
    \begin{axis}[
        axis on top=true,
        axis x line=middle,
        axis y line=middle,
        ymin=-0.5, ymax=4.5,
        ytick={0.5,1.0,2.0,2.5,3.25,4.25},
        yticklabels={$a_1$, $a_2$, $a_j$, $a_{j+1}$, $a_{m-1}$, $a_m$},
        xmin=-0.5, xmax=7,
        xtick={1.25,2.0,3.25,4.0,5.5,6.25},
        tick style={line width=0.25, gray},
        tick label style={minimum width=3em, minimum height=2em},
        xticklabels={$b_1$, $b_2$, $b_j$, $b_{j+1}$, $b_{m-1}$, $b_m$},
        extra x ticks={2.65, 4.75},
        extra x tick labels={$\dots$, $\dots$},
        extra x tick style={tick style={draw=none}},
        extra y ticks={1.575, 2.975},
        extra y tick labels={$\vdots$, $\vdots$},
        extra y tick style={tick style={draw=none}},
    ]
    
    \draw[line width=0.25, gray] (0, 0.5) -- (1.25, 0.5);
    \draw[line width=0.25, gray] (0, 1.0) -- (2.0, 1.0);
    \draw[line width=0.25, gray] (0, 2.0) -- (3.25, 2.0);
    \draw[line width=0.25, gray] (0, 2.5) -- (4.0, 2.5);
    \draw[line width=0.25, gray] (0, 3.25) -- (5.5, 3.25);
    \draw[line width=0.25, gray] (0, 4.25) -- (6.25, 4.25);
    
    \draw[line width=0.25, gray] (1.25, 0) -- (1.25, 0.5);
    \draw[line width=0.25, gray] (2.0, 0) -- (2.0, 1.0);
    \draw[line width=0.25, gray] (3.25, 0) -- (3.25, 2.0);
    \draw[line width=0.25, gray] (4.0, 0) -- (4.0, 2.5);
    \draw[line width=0.25, gray] (5.5, 0) -- (5.5, 3.25);
    \draw[line width=0.25, gray] (6.25, 0) -- (6.25, 4.25);
    
    \draw[color0, thick] (0.0, 0.0) -- (1.25, 0.5) -- (2.0, 1.0);
    \draw[color0, thick, dashed] (2.0, 1.0) -- (3.25, 2.0);
    \draw[color0, thick] (3.25, 2.0) -- (4.0, 2.5);
    \draw[color0, thick, dashed] (4.0, 2.5) -- (5.5, 3.25);
    \draw[color0, thick] (5.5, 3.25) -- (6.25, 4.25);

    \end{axis}
\end{tikzpicture}
\end{tabular}
     }
     \caption{The function $f_{j+1}(x)$ transforms the points $0=a_0=b^j_0, a_1=b^j_1,\dots,a_j=b^j_j$ and $b^j_{j+1},\dots,b^j_m$ (left). The region $x\leq a_j$ is left unchanged and the region $x\geq a_j$ is linearly rescaled so that $f_{j+1}(b^j_{j+1})=a_{j+1}$. The piecewise-linear function $f=f_m\circ\dots\circ f_0$ maps $b_j$ to $a_j$ (right).}
     \label{fig:mapping}
\end{figure}

Next we derive the ReLU layers $\skl{F_j}_j$ from the functions $\skl{f_j}_j$. For $j=0,\dots,m-1$ we denote
\[
 w_{j+1} = \frac{a_{j+1} - b^j_{j+1}}{b^j_{j+1} - a_j}
\]
and rewrite \eqref{eq:fj_resize} as
\[
 f_{j+1}(x) = \bmat{1 & w_{j+1}}\varrho\kl{ \bmat{1 \\ 1} x + \bmat{0 \\ -a_j}}\quad\text{for}\quad x\geq 0.
\]
Two consecutive maps can be combined as
\[
    f_{j+1}\kl{f_{j}(x)} = \bmat{1 & w_{j+1}} \varrho \kl{ \bmat{1 & w_{j} \\ 1 & w_{j}} \varrho \kl{\bmat{1 \\1}x + \bmat{0 \\ -a_{j-1}}   } + \bmat{0 \\ -a_{j}}  }
\]
for $x\geq 0$, hence we set
\[
 F_{j}\colon\R^2 \to \R^2 \colon\bfx\mapsto
\varrho \kl{ \bmat{1 & w_{j} \\ 1 & w_{j}} \bfx + \bmat{0 \\ -a_{j}}  }
\]
for $j=1,...,m-1$. It is now easy to check that with
\[
 F_0(\bfx) = \varrho\kl{\bmat{ 1 & 0 \\ 1 & 0} \bfx }
\]
and
\[
  F_m(\bfx) \mapsto \varrho\kl{ \bmat{1 & w_m \\ 0 & 0} \bfx }
\]
we finally get
\[
 F\kl{\bmat{x \\ 0}} = \bmat{f\kl{x} \\ 0}\quad\text{for}\quad x\geq 0
\]
for $F=F_m\circ F_{m-1} \circ \dots \circ F_1 \circ F_0$.
\end{proof}

\subsection{Bending the Arc}

\begin{figure}
     \centering
     \resizebox{\textwidth}{!}{
     % \scriptsize

\pgfmathsetmacro{\phi}{90/3}
\pgfmathsetmacro{\a}{0.5}
\pgfmathsetmacro{\b}{0.3}
\pgfmathsetmacro{\c}{0.25}

\pgfmathsetmacro{\o}{1.4}
\pgfmathsetmacro{\u}{-0.2}

\begin{tabular}{@{}l@{}c@{}l@{}c@{}l@{}c@{}l@{}c@{}}
i) &
    \raisebox{-.9\height}{\begin{tikzpicture}[scale=2.5]
        \draw[thin, gray, ->, >=latex] (0, \u) -- (0, \o);
        \draw[thin, gray, ->, >=latex] (-0.2, 0) -- (0.6, 0);

        \polychain[color0, thick]({0,0,0}, {\a,\b,\c}, (0,0));

        \node[above right] at (co0) {$\bfv^0_0$};
        \node[right] at (co1) {$\bfv^0_1$};
        \node[right] at (co2) {$\bfv^0_2$};
        \node[right] at (co3) {$\bfv^0_3$};
    \end{tikzpicture}}
& ii) &
    \raisebox{-.9\height}{\begin{tikzpicture}[scale=2.5]
        \draw[thin, gray, ->, >=latex] (0, \u) -- (0, \o);
        \draw[thin, gray, ->, >=latex] (-0.5, 0) -- (0.6, 0);

        \pgfmathsetmacro{\offset}{-cos(90 - \phi)*(\a+\b)};
        \polychain[gray]({\phi,0,0}, {\a,\b,\c}, (\offset,0), 4);

        \node[above left] at (co0) {$\bfw^0_0$};
        \node[left] at (co1) {$\bfw^0_1$};

        \pgfmathsetmacro{\a}{cos(\phi)*\a};
        \pgfmathsetmacro{\b}{cos(\phi)*\b};

        \polychain[color0, thick]({0,0,\phi}, {\a,\b,\c}, (0,0), 4);
        \node[above right] at (co0) {$\bfv^1_0$};
        \node[right] at (co1) {$\bfv^1_1$};
        \node[right] at (co2) {$\bfv^1_2$};
        \node[right] at (co3) {$\bfv^1_3$};

    \end{tikzpicture}}
& iii) &
    \raisebox{-.9\height}{\begin{tikzpicture}[scale=2.5]
        \draw[thin, gray, ->, >=latex] (0, \u) -- (0, \o);
        \draw[thin, gray, ->, >=latex] (-0.3, 0) -- (0.6, 0);

        \pgfmathsetmacro{\offset}{-cos(90 - \phi)*(\a)};
        \polychain[gray]({\phi,0,\phi}, {\a,\b,\c}, (\offset,0), {1,1,1,1});

        \node[above left] at (co0) {$\bfw^1_0$};

        \pgfmathsetmacro{\a}{cos(\phi)*\a};
        \polychain[color0, thick]({0,\phi,\phi}, {\a,\b,\c}, (0,0), {1,1,1});
        \node[above right] at (co0) {$\bfv^2_0$};
        \node[right] at (co1) {$\bfv^2_1$};
        \node[above] at (co2) {$\bfv^2_2$};
        \node[above] at (co3) {$\bfv^2_3$};
    \end{tikzpicture}}
& iv) &
    \raisebox{-.9\height}{\begin{tikzpicture}[scale=2.5]
        \draw[thin, gray, ->, >=latex] (0, \u) -- (0, \o);
        \draw[thin, gray, ->, >=latex] (-0.2, 0) -- (0.9, 0);

        \polychain[color0, thick]({\phi,\phi,\phi}, {\a,\b,\c}, (0,0), {1,1,1,1});
        \node[above left] at (co0) {$\bfv^3_0$};
        \node[above] at (co1) {$\bfv^3_1$};
        \node[above] at (co2) {$\bfv^3_2$};
        \node[above] at (co3) {$\bfv^3_3$};
    \end{tikzpicture}}
\end{tabular}
     }
     \caption{Illustration of the 'bending' of a 3-arc in 4 steps. Step (i) shows $G_0$ ($90^{\circ}$ rotation) applied to $[a_m,a_0]\times\{0\}$.  Steps (ii) and (iii) show the image of $G_1\circ G_0$ and $G_2\circ G_1\circ G_0$ respectively (blue) as well as the corresponding transform just before the last ReLU-activation (gray). Step (iv) shows the final arc arising as the image of $G=G_3\circ G_2\circ G_1\circ G_0$.
     }
     \label{fig:bending}
\end{figure}
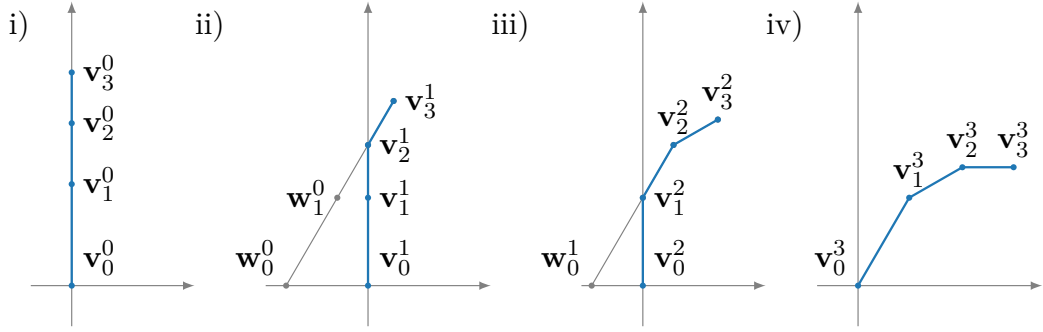

 Let us next consider how to ``bend'' a one dimensional subspace into an $m$-arc. This iterative procedure is illustrated in \cref{fig:bending}. It starts with all vertices in a straight line on the $x_2$-axis. First the polygonal chain is rotated by $-\phi_m$ (``to the right'') and translated in negative $x_1$ direction (``to the left'') such that the second last vertex lies on the $x_2$-axis. Then a ReLU is applied, which projects all vertices up until the second last onto the $x_2$-axis. The last last arc segment now has the correct angle and the process is repeated until all segments are correctly ``bent''. Note that each projection shrinks the unbent line segments by a factor of $\cos(\phi_m)$.
 To describe this transformation formally, we denote rotation matrices in $\R^2$ by $$R_\alpha=\begin{bmatrix}\cos(\alpha) & -\sin(\alpha) \\ \sin(\alpha) & \cos(\alpha) \end{bmatrix}.$$
 Furthermore, for $m\geq 2$ and $0=a_0<a_1<\dots<a_m<\infty$ we define
 \[
  \bfd_j = \bmat{\sin\kl{\phi_m}\cos\kl{\phi_m}^j a_{m-j} \\ 0},\quad\text{for}\quad j=1,\dots,m,
 \]
 as well as
 \begin{align*}
     &G_0\colon \R^2 \rightarrow \R^2\colon \bfx \mapsto \varrho\kl{R_{\frac{\pi}{2}}\bfx}
     \intertext{and}
     &G_j\colon \R^2 \rightarrow \R^2\colon \bfx \mapsto \varrho\kl{R_{-\phi_m}\bfx - \bfd_j} \quad\text{for}\quad j=1,\dots,m,
 \end{align*}
 which clearly are ReLU-transformations.

  \begin{lemma}\label{lem:nn_relu_a2arc}
 Let $m\geq 2$ and $0=a_0<a_1<\dots<a_m<\infty$ and $G_j$ as above. Then
 \[
    G\colon[a_0,a_m]\times\{0\}\subseteq\R^2\to\R^2\colon \bfx\mapsto (G_m \circ G _{m-1}\circ \dots\circ G_1\circ G_0)(\bfx)
 \]
 is a parametrisation of a standard $m$-arc with vertices $\bfv_i=G\left(\begin{bmatrix}a_i \\ 0\end{bmatrix}\right)$, line segments $\ell_i=G([a_{i-1},a_i]\times\{0\})$, and scaling factors $r_i = \cos^{m-i}(\phi_m)(a_i-a_{i-1})$ for $i=1,\dots,m$.
 \end{lemma}

 \begin{proof}
  We need to show that $G$ is (in its first component) a piecewise-linear function with breakpoints $a_i$, and that $\bfv_0 = \bfzero$ as well as
  \begin{equation}\label{eq:bend_claim}
      \bfv_i = \bfv_{i-1} + \cos^{m-i}(\phi_m)(a_i-a_{i-1}) \bmat{\sin(i\phi_m) \\ \cos(i\phi_m)},\quad\text{for}\quad i=1,\dots,m.
  \end{equation}
  For this we denote \[\bfv_i^j = (G_j\circ\dots\circ G_0)\kl{\bmat{a_i \\ 0}}\quad\text{for}\quad i,j=0,...,m.\] By the choice of $\phi_m$ we know $\sin(\phi_m)>0$ and $\cos(\phi_m)>0$. Hence, the first component of the bias vectors $\bfd_j$ satisfy $\sin(\phi_m)\cos(\phi_m)^{j}a_{m-j}\geq 0$ for $j=1,\dots,m$.
  Therefore, $\bfv^j_0=\bfzero$ for all $j$ and in particular $\bfv_0=\bfv^m_0=\bfzero$. Further, we will show inductively over $j$ that for any $i,j$ we have
  \begin{equation}\label{eq:bend_induction}
   \bfv^j_i-\bfv^j_{i-1} = \begin{cases}
   \cos\kl{\phi_m}^{j} (a_i-a_{i-1}) \bmat{0 \\ 1}, &\quad i\leq m - j, \\
   \cos\kl{\phi_m}^{m-i} (a_i-a_{i-1}) \bmat{\sin\kl{(i+j-m)\phi_m} \\ \cos\kl{(i+j-m)\phi_m}}, & \quad i> m-j.
   \end{cases}
\end{equation}
  Then \eqref{eq:bend_claim} follows from \eqref{eq:bend_induction} with $j=m$.

  To start the inductive proof we observe that for $j=0$ and $i=1,\dots,m$ we have
  \begin{align*}
     \bfv_i^0 - \bfv_{i-1}^0 &= G_0\kl{\bmat{a_i \\ 0}} - G_0\kl{\bmat{a_{i-1} \\ 0}} \\
     &= R_{\frac{\pi}{2}}\bmat{a_i \\ 0} - R_{\frac{\pi}{2}}\bmat{a_{i-1} \\ 0} \\
     &= (a_i-a_{i-1})\bmat{0\\1},
  \end{align*}
  satisfying \eqref{eq:bend_induction}.

  Next, let us assume that \eqref{eq:bend_induction} holds for some $j<m$. For any $i$ we denote the vectors
  $\bfw^j_i = R_{-\phi_m}\bfv^j_i - \bfd_j$ and therefore $\bfv^{j+1}_i = G_{j+1}(\bfv^j_i) = \varrho(\bfw^j_i)$ and
  \begin{equation*}\label{eq:bend_pre_relu}
    \bfw^j_i - \bfw^j_{i-1} =
    \begin{cases}
        \cos\kl{\phi_m}^{j} (a_i-a_{i-1}) \bmat{\sin\kl{\phi_m} \\ \cos\kl{\phi_m}}, &\enskip i\leq m - j, \\
        \cos\kl{\phi_m}^{m-i} (a_i-a_{i-1}) \bmat{\sin\kl{(i+j+1-m)\phi_m} \\ \cos\kl{(i+j+1-m)\phi_m}}, &\enskip i> m-j.
   \end{cases}
 \end{equation*}
 We also have $\sin((i+j+1-m)\phi_m)> 0$ and $\cos((i+j+1-m)\phi_m)\geq 0$ for $i>m-j$ by the choice of $\phi_m$ and therefore $\bfw^j_m\geq \bfw^j_{m-1}\geq\dots\geq\bfw^j_1\geq\bfw^j_0=-\bfd_j$ component-wise, where even all inequalities are strict in the first component. We immediatley see that all $\bfw^j_i$ lie in the upper half-plane. Further,
 \begin{align*}
  \bfw^j_{m-j} &= \bfw_0^j + \sum_{i=1}^{m-j} (\bfw^j_i - \bfw^j_{i-1})\\
   &= -\bfd_j + \sum_{i=1}^{m-j}\cos\kl{\phi_m}^{j}(a_i-a_{i-1}) \bmat{\sin\kl{\phi_m} \\ \cos\kl{\phi_m}} \\
   &= -\bmat{\sin\kl{\phi_m}\cos\kl{\phi_m}^j a_{m-j} \\ 0}  +
   \cos\kl{\phi_m}^{j} a_{m-j} \bmat{\sin\kl{\phi_m} \\ \cos\kl{\phi_m}} \\
   &= \cos\kl{\phi_m}^{j+1} a_{m-j} \bmat{0 \\ 1},
 \end{align*}
 from which we can conclude that
 \begin{align}
     (\bfw_i^j)_1 &\geq 0,\quad\text{for}\quad i=1,\dots,m \label{eq:bending1}\\
     (\bfw_i^j)_2 &< 0,\quad\text{for}\quad i=1,...,m-j-1, \label{eq:bending2}\\
     (\bfw_i^j)_2 &\geq 0,\quad\text{for}\quad i=m-j,...,m. \label{eq:bending3}
 \end{align}
 Now $\bfv^{j+1}_i = \varrho(\bfw^j_i)$ together with \eqref{eq:bending1} to \eqref{eq:bending3} implies
  \begin{alignat*}{2}
     (\bfv_i^{j+1})_1 &= (\bfw_i^j)_1,&&\quad\text{for}\quad i=1,...,m, \\
     (\bfv_i^{j+1})_2 &= 0,&&\quad\text{for}\quad i=1,...,m-j-1, \\
     (\bfv_i^{j+1})_2 &= (\bfw_i^j)_2,&&\quad\text{for}\quad i=m-j,...,m,
 \end{alignat*}
 and therefore $\bfv^{j+1}_i-\bfv^{j+1}_{i-1}$ is equivalent to
 \[
   \begin{cases}
   \cos\kl{\phi_m}^{j+1} (a_i-a_{i-1}) \bmat{0 \\ 1}, &\enskip i\leq m - (j+1) \\
   \cos\kl{\phi_m}^{m-i} (a_i-a_{i-1}) \bmat{\sin\kl{(i+j+1-m)\phi_m} \\ \cos\kl{(i+j+1-m)\phi_m}}, &\enskip i> m-(j+1),
   \end{cases}
  \]
  which shows that \eqref{eq:bend_induction} holds for $j+1$ and concludes the inductive step.

  Each $G_{j+1}$ adds only one new breakpoint at to the overall function, corresponding to point $\bfw^j_{m-j}$ where the second component of $(G_j\circ\dots\circ G_0)(\bfx)$ switches sign. This corresponds to the breakpoint $a_{m-j}$ of $G$.
 \end{proof}

 We are now ready to given the main proof of this section.
 \begin{proof}[Proof of \cref{lem:xi_surjective}]
 Let $m\in\N$ be arbitrary. By assumption there exists a measure $\mu\in\{p(\theta)\}_{\theta\in\Omega}$ with infinite support. As discussed in \cref{sec:onedim_twodim} and in particular by using \cref{lem:proj_one_dim} we can without loss of generality assume $d=2$ and $\supp(\mu)$ is compact and contained in the non-negative part of $\linspan\{\bfe_1\}$. In other words we can rewrite $\supp(\mu)=S\times\{0\}\subseteq[0,b]\times\{0\}$ for an infinite set $S\subseteq\R_{\geq 0}$ and some $b>0$. By \cref{lem:interval_partition} we can choose points $0=b_0<b_1<\dots<b_m=b$ so that $S$ intersects each $\kl{b_{j-1}, b_j}$. By the choice of $S$ and since $\kl{b_{j-1},b_j}\times\{0\}$ is relatively open in $S\times\{0\}$ we have $\mu((b_{j-1},b_j)\times\{0\})>0$ for all $j$. Set $0<\delta\leq\min_j \mu((b_{j-1},b_j)\times\{0\})$.

 Now let $A\in\CA_m$ be an arbitrary standard $m$-arc and denote its set of vertices by $(\bfv_0,\dots,\bfv_m)=\vertex(A)$, its line segments by $(\ell_1,\dots,\ell_m)=\segment(A)$, and its scaling factors by $(r_1,\dots,r_m)=\scale(A)$. We set $a_0=0$ and iteratively $a_j=a_{j-1} + r_j \cos^{j-m}(\phi_m)$ for $j=1,\dots,m$. By \cref{lem:nn_relu_b2a} there is a ReLU neural network $F$ satisfying
 \[
 F\kl{\begin{bmatrix}b_j \\ 0\end{bmatrix}} = \begin{bmatrix}a_j \\ 0\end{bmatrix}\quad\text{for}\quad j=1,\dots,m,
 \]
 and then by \cref{lem:nn_relu_a2arc} another ReLU neural network $G$ satisfying
 \[
 G\kl{\begin{bmatrix}a_i \\ 0\end{bmatrix}} = \bfv_j\quad\text{for}\quad j=1,\dots,m.
 \]
 Altogether $G\circ F$ is a ReLU neural network transforming $\bfb_j=\begin{bmatrix}b_i & 0\end{bmatrix}^\top$ to the vertex $\bfv_j$ and $\ekl{b_{j-1},b_j}\times\{0\}$ to the line segment $\ell_j$.
 By \cref{lem:cont_supp} we have
 \[
    \supp\kl{(G\circ F)_\ast\mu} = \overline{(G\circ F)(\supp(\mu))} \subseteq \overline{(G\circ F)(\ekl{b_0,b_m}\times\{0\})} \subseteq A.
 \]
 Also, for each $j$ we get
 \[
 ((G\circ F)_\ast\mu)(\ell_j)=\mu\kl{(G\circ F)^{-1}(\ell_j)} \geq \mu(\kl{b_{j-1},b_j}\times\{0\})\geq\delta
 \]
 and therefore $(G\circ F)_\ast \mu \in\CD_{\delta,m}(\R^2)$ and $\arc\kl{(G\circ F)_\ast \mu}=A$.
 Using the ReLU-invariance, we know that $(G\circ F)_\ast\mu\in \{p(\theta)\}_{\theta\in\Omega_{\delta,m}}\subseteq \{p(\theta)\}_{\theta\in\Omega}$. Altogether, since $A\in\CA_m$ was arbitrary we conclude that $\arc\circ\,p\vert_{\Omega_{\delta,m}}$ is surjective onto $\CA_m$. Clearly, the map $\scale\colon\CA_m\to\R^{m-1}_+\times\{1\}$ is a bijection, hence also $\Xi\colon\Omega_{\delta,m}\to\R^{m-1}_+\times\{1\}$ is surjective.
 \end{proof}

 \section{Proof of \cref{lem:xi_loclip}: Local Lipschitz Continuity of \texorpdfstring{$\Xi$}{Xi}}\label{sec:xi_loclip}
 We will now show that the map $\Xi\colon\Omega_{\delta,m}\to\R^m_+$ is locally Lipschitz continuous by showing this for each of its three composite parts $p\vert_{\Omega_{\delta,m}}\colon\Omega_{\delta,m}\to\CD_{\delta,m}(\R^d)$, $\arc\colon\CD_{\delta,m}(\R^d)\to\CA_m$, and $\scale\colon\CA_m\to\R^m_+$.

 Firstly, $p\vert_{\Omega_{\delta,m}}\colon\Omega_{\delta,m}\to\CD_{\delta,m}(\R^d)$ is locally Lipschitz continuous by \cref{asm:main}, so there is nothing to show.

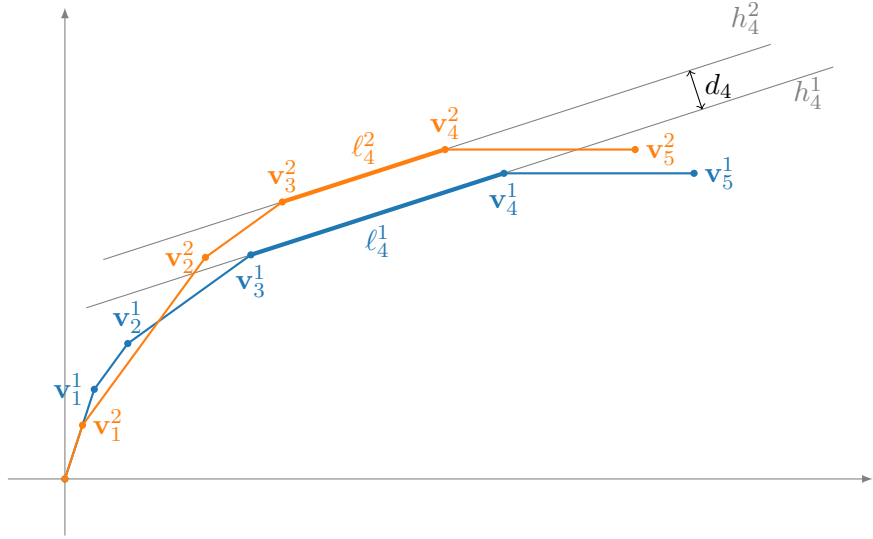
\begin{figure}
    \centering
    % \scriptsize

\begin{tikzpicture}[scale=2.5]

\draw[thin, gray, ->, >=latex] (0, -0.3) -- (0, 2.5);
\draw[thin, gray, ->, >=latex] (-0.3, 0) -- (4.25, 0);

\coordinate (v0) at (0.0, 0.0);

\coordinate (v11) at ($(v0)+(90-1*90/5:0.5)$);
\coordinate (v12) at ($(v11)+(90-2*90/5:0.3)$);
\coordinate (v13) at ($(v12)+(90-3*90/5:0.8)$);
\coordinate (v14) at ($(v13)+(90-4*90/5:1.4)$);
\coordinate (v15) at ($(v14)+(90-5*90/5:1.0)$);

\coordinate (v21) at ($(v0)+(90-1*90/5:0.3)$);
\coordinate (v22) at ($(v21)+(90-2*90/5:1.1)$);
\coordinate (v23) at ($(v22)+(90-3*90/5:0.5)$);
\coordinate (v24) at ($(v23)+(90-4*90/5:0.9)$);
\coordinate (v25) at ($(v24)+(90-5*90/5:1.0)$);

\draw[add=0.65 and 1.3, thin, gray] (v13) to (v14) node[below left] {$h_4^1$};
\draw[add=1.1 and 2, thin, gray] (v23) to (v24)  node[above left] {$h_4^2$};
\draw[ultra thick, color0] (v13) -- (v14) node[midway, below] {$\ell_4^1$};
\draw[ultra thick, color1] (v23) -- (v24) node[midway, above] {$\ell_4^2$};

\coordinate (P) at ($(v23)+2.5*(v24)-2.5*(v23)$);
\draw[<->] (P) -- ($(v13)!(P)!(v14)$) node[pos=.4, right] {$d_4$};

\draw (v11) node[left, color0] {$\bfv_1^1$};
\draw (v12) node[above, color0] {$\bfv_2^1$};
\draw (v13) node[below, color0] {$\bfv_3^1$};
\draw (v14) node[below, color0] {$\bfv_4^1$};
\draw (v15) node[right, color0] {$\bfv_5^1$};

\draw (v21) node[right, color1] {$\bfv_1^2$};
\draw (v22) node[left, color1] {$\bfv_2^2$};
\draw (v23) node[above, color1] {$\bfv_3^2$};
\draw (v24) node[above, color1] {$\bfv_4^2$};
\draw (v25) node[right, color1] {$\bfv_5^2$};

\marc[color0, thick](5, {0.5, 0.3, 0.8, 1.4, 1.0});
\marc[color1, thick](5, {0.3, 1.1, 0.5, 0.9, 1.0});

\end{tikzpicture}
    \caption{Example of two standard $m$-arcs $A_1$ (blue) and $A_2$ (orange) for $m=5$. Supporting affine subspaces $h_4^1$ and $h_4^2$ and their distance $d_4$ are shown exemplarily for the fourth line segments. At this segment $A_2$ is the outer arc and $A_1$ is the inner arc.}
    \label{fig:outer-arc}
 \end{figure}

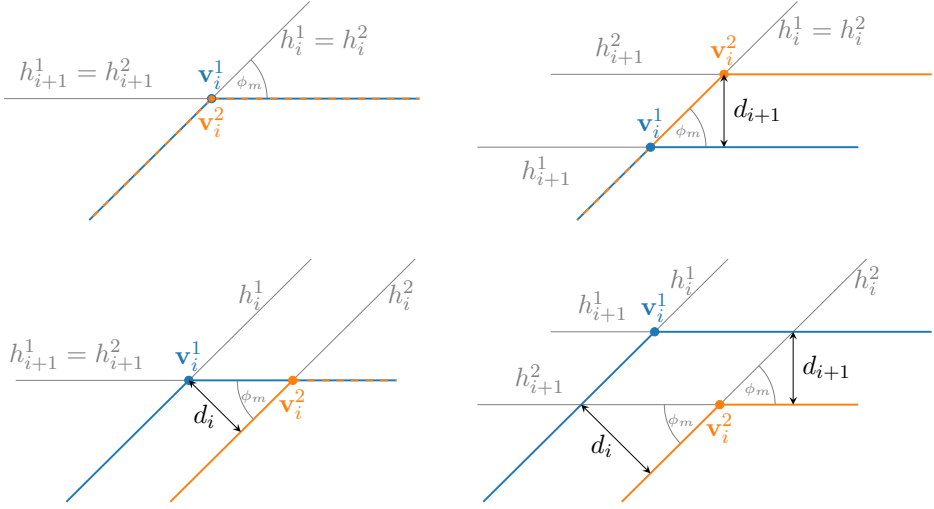
\begin{figure}
     \centering
     \resizebox{0.85\textwidth}{!}{
     % \scriptsize
\begin{tabular}{@{}c@{\qquad}c@{}}
\begin{tikzpicture}[scale=2]
\coordinate (vi1) at (0.0, 0.0);
\coordinate (vi2) at (vi1);

\draw[gray, thin] (vi1) -- ++ (45:1) node[pos=0.6, right]{$h_i^1=h_i^2$};
\draw[gray, thin] (vi1) -- ++ (0:-1.5) node[pos=0.6, above]{$h_{i+1}^1=h_{i+1}^2$};
\draw[gray, thin] (vi1)++(0:0.4) arc (0:45:0.4);
\node[gray] at ($(vi1)+(22.5:0.3)$) {\tiny $\phi_m$};

\draw[color0, thick] (vi1) -- ++ (45:-1.25);
\draw[color0, thick] (vi1) -- ++ (0:1.5);

\draw[color1, thick, dashed] (vi2) -- ++ (45:-1.25);
\draw[color1, thick, dashed] (vi2) -- ++ (0:1.5);

\draw[fill, color0] (vi1)  circle[radius=0.03] node[above]{$\bfv_i^1$};
\draw[fill, color1] (vi2)  circle[radius=0.02] node[below]{$\bfv_i^2$};
\end{tikzpicture}
&
\begin{tikzpicture}[scale=2]
\coordinate (vi1) at (0.0, 0.0);
\coordinate (vi2) at ($(vi1)+(45:0.75)$);

\draw[gray, thin] (vi2) -- ++ (45:0.75) node[pos=0.6, right]{$h_i^1=h_i^2$};
\draw[gray, thin] (vi1) -- ++ (0:-1.25) node[pos=0.6, below]{$h_{i+1}^1$};
\draw[gray, thin] (vi2) -- ++ (0:-1.25) node[pos=0.6, above]{$h_{i+1}^2$};
\draw[gray, thin] (vi1)++(0:0.4) arc (0:45:0.4);
\node[gray] at ($(vi1)+(22.5:0.3)$) {\tiny $\phi_m$};

\draw[color0, thick] (vi1) -- ++ (45:-0.75);
\draw[color0, thick] (vi1) -- ++ (0:1.5);

\draw[color1, thick, dashed] (vi1) -- ++ (45:-0.75);
\draw[color1, thick] (vi2) -- (vi1);
\draw[color1, thick] (vi2) -- ++ (0:1.5);

\draw[fill, color0] (vi1)  circle[radius=0.03] node[above]{$\bfv_i^1$};
\draw[fill, color1] (vi2)  circle[radius=0.03] node[above]{$\bfv_i^2$};

\draw[<->, >=stealth] (vi2) -- (vi1-|vi2) node[midway, right] {$d_{i+1}$};
\end{tikzpicture}
\\[1em]
\begin{tikzpicture}[scale=2]
\coordinate (vi1) at (0.0, 0.0);
\coordinate (vi2) at ($(vi1)+(0:0.75)$);

\draw[gray, thin] (vi1) -- ++ (45:1.25) node[pos=0.7, left]{$h_i^1$};
\draw[gray, thin] (vi2) -- ++ (45:1.25) node[pos=0.7, right]{$h_i^2$};
\draw[gray, thin] (vi1) -- ++ (0:-1.25) node[pos=0.2, above left]{$h_{i+1}^1=h_{i+1}^2$};
\draw[gray, thin] (vi2)++(0:-0.4) arc (0:45:-0.4);
\node[gray] at ($(vi2)+(22.5:-0.3)$) {\tiny $\phi_m$};

\draw[color0, thick] (vi1) -- ++ (45:-1.25);
\draw[color0, thick] (vi1) -- ++ (0:1.5);

\draw[color1, thick] (vi2) -- ++ (45:-1.25);
\draw[color1, thick, dashed] (vi2) -- ++ (0:0.75);

\draw[fill, color0] (vi1)  circle[radius=0.03] node[above]{$\bfv_i^1$};
\draw[fill, color1] (vi2)  circle[radius=0.03] node[below]{$\bfv_i^2$};

\draw[<->, >=stealth] (vi1) -- ($(vi2)!(vi1)!($(vi2)+(45:1)$)$) node[pos=0.3, below] {$d_i$};
\end{tikzpicture}
&
\begin{tikzpicture}[scale=2]
\coordinate (vi1) at (0.0, 0.0);
\coordinate (vi2) at ($(vi1)+(0:1.0)+(45:-0.75)$);

\draw[gray, thin] (vi1) -- ++ (45:0.75) node[pos=0.7, left]{$h_i^1$};
\draw[gray, thin] (vi1) -- ++ (0:-0.75) node[pos=0.5, above]{$h_{i+1}^1$};
\draw[gray, thin] (vi2) -- ++ (45:1.5) node[pos=0.85, right]{$h_i^2$};
\draw[gray, thin] (vi2) -- ++ (0:-1.75) node[pos=0.6, above left]{$h_{i+1}^2$};
\draw[gray, thin] (vi2)++(0:0.4) arc (0:45:0.4);
\node[gray] at ($(vi2)+(22.5:0.3)$) {\tiny $\phi_m$};
\draw[gray, thin] (vi2)++(0:-0.4) arc (0:45:-0.4);
\node[gray] at ($(vi2)+(22.5:-0.3)$) {\tiny $\phi_m$};

\draw[color0, thick] (vi1) -- ++ (45:-1.75);
\draw[color0, thick] (vi1) -- ++ (0:2);

\draw[color1, thick] (vi2) -- ++ (45:-1);
\draw[color1, thick] (vi2) -- ++ (0:1);

\draw[fill, color0] (vi1)  circle[radius=0.03] node[above]{$\bfv_i^1$};
\draw[fill, color1] (vi2)  circle[radius=0.03] node[below]{$\bfv_i^2$};

\draw[<->, >=stealth] ($(vi1)+(0:1.0)$) -- ($(vi2)!($(vi1)+(0:1.0)$)!($(vi2)+(0:1.0)$)$) node[pos=0.5, right] {$d_{i+1}$};
\draw[<->, >=stealth] ($(vi2)+(0:-1.0)$) -- ($(vi2)!($(vi2)+(0:-1.0)$)!($(vi2)+(45:0.75)$)$) node[pos=0.3, below] {$d_{i}$};
\end{tikzpicture}

\end{tabular}
     }
     \caption{Four possible arrangements of corresponding vertices $\bfv_i^1$ and $\bfv_i^2$ of two arcs. In case one both vertices are equal (top left). In cases two and three the vertices are different but either the affine subspaces $h_i^1$ and $h_i^2$ or $h_{i+1}^1$ and $h_{i+1}^2$ coincide (top right and bottom left). In the fourth case all affine subspaces differ and their intersections form a parallelogram with $\bfv_i^1$ and $\bfv_i^2$ at opposite corners (bottom right).}
     \label{fig:vertex-arrangements}
\end{figure}

 Secondly, the local Lipschitz continuity of $\arc\colon\CD_{\delta,m}(\R^d)\to\CA_m$ can be derived from geometric observations.
 \begin{lemma}\label{lem:loclip1}
    The map $\arc\colon(\CD_{\delta,m}(\R^d),d_P)\to(\CA_m,d_\CA)\colon \mu\mapsto A_\mu$, is locally Lipschitz continuous with Lipschitz constant $\frac{2\sqrt{2}}{\sin\kl{\phi_m}}$.
 \end{lemma}
 \begin{proof}
 Let $\mu_0\in\CD_{\delta, m}(\R^d)$ be arbitrary. We will show Lipschitz continuity on the open ball of radius $\frac{\delta}{2}$ around $\mu_0$. For this let $\mu_1,\mu_2\in B_{\frac{\delta}{2}}(\mu_0) \cap \CD_{\delta, m}(\R^d)$ and therefore $d_P(\mu_1,\mu_2)\leq \delta$. We know that $\mu_1$ and $\mu_2$ are $\delta$-supported on two unique standard $m$-arcs $A_1=\arc\kl{\mu_1}=A_{\mu_1}\in\CA_m$ and $A_2=\arc\kl{\mu_2}=A_{\mu_2}\in\CA_m$ and by definition these lie in the same two-dimensional subspace $\linspan\skl{\bfe_1,\bfe_2}$ of $\R^d$. So without loss of generality we carry out the remaining proof in $\R^2$. Let $(\ell_1^1,\dots,\ell_m^1) = \segment(A_1)$ and $(\ell_1^2,\dots,\ell_m^2) = \segment(A_2)$ denote the line segments of $A_1$ and $A_2$ respectively. We denote the affine hulls of individual line segments as $h_i^j = \affspan\kl{\ell_i^j}$ and further denote the Euclidean distance of two corresponding affine hulls by $d_i=\dist(h_i^1, h_i^2)$. An example for this can be seen in \cref{fig:outer-arc}.

 We want to upper bound the Euclidean distances $\|\bfv_i^1-\bfv_i^2\|_2$ of all corresponding vertices of the two arcs. For any $i$ one of four cases can occur, as visualised in \cref{fig:vertex-arrangements}. Firstly, if $\bfv_i^1=\bfv_i^2$ (cf.\ \Cref{fig:vertex-arrangements}, top left) we trivially get $\|\bfv_i^1-\bfv_i^2\|_2=0$.
 Secondly, if $\bfv_i^1\neq\bfv_i^2$ but $h_i^1=h_i^2$ (cf.\ \Cref{fig:vertex-arrangements}, top right) it is not hard to see that
 \[
    \|\bfv_i^1-\bfv_i^2\|_2 \leq \frac{d_{i+1}}{\sin\kl{\phi_m}}.
 \]
 Thirdly, if $\bfv_i^1\neq\bfv_i^2$ but $h_{i+1}^1=h_{i+1}^2$ (cf.\ \Cref{fig:vertex-arrangements}, bottom left) we similarly get
 \[
    \|\bfv_i^1-\bfv_i^2\|_2 \leq \frac{d_{i}}{\sin\kl{\phi_m}}.
 \]
 In the fourth case, where $\bfv_i^1\neq\bfv_i^2$, $h_i^1\neq h_i^2$, and $h_{i+1}^1\neq h_{i+1}^2$ the four intersections of the affine spaces form a parallelogram (cf.\ \Cref{fig:vertex-arrangements}, bottom right). We obtain
 \[
    \|\bfv_i^1-\bfv_i^2\|_2 \leq \frac{\sqrt{2}\kl{d_{i}+d_{i+1}}}{\sin\kl{\phi_m}},
 \]
 using the parallelogram identity. Thus, it remains to bound all the distances $d_i$.

 For this, assume that for some $i$ we have $d_i>0$. We say that $\ell_i^1$ is the \emph{outer} line segment and $\ell_i^2$ is the \emph{inner} line segment if $\dist\kl{\bfzero, h_i^1} > \dist\kl{\bfzero, h_i^2}$ and vice versa if $\dist\kl{\bfzero, h_i^1} < \dist\kl{\bfzero, h_i^2}$. Without loss of generality assume that $\ell_i^1$ is the outer line segment. Then for any point $\bfp\in A_2$ we have $\dist(\bfp, \ell_i^1) \geq d_i$. Thus $\mu_2((\ell_i^1)^\epsilon)=0$ for any $\epsilon<d_i$. But $\mu_1(\ell_i^1)\geq \delta$ and therefore by definition of the Prokhorov metric
 \[
  d_i \leq d_P(\mu_1,\mu_2).
 \]
 Altogether, we obtain
 \[
    d_\CA\kl{A_1, A_2} = \max_{i} \|\bfv_i^1-\bfv_i^2\|_2 \leq \frac{2 \sqrt{2}}{\sin\kl{\phi_m}} d_P\kl{\mu_1,\mu_2}. \qedhere
 \]
 \end{proof}

 Thirdly, the local Lipschitz continuity of $\scale\colon\CA_m\to\R^m_+$ is a straight-forward calculation if $\R^m_+$ is equipped with the metric induced by the $\ell_\infty$-norm. By norm equivalence the same also holds for $\R^m_+$ viewed as a subspace of Euclidean $\R^m$.
 \begin{lemma}\label{lem:loclip2}
    The map $\scale\colon(\CA_m, d_\CA)\to(\R_+^m,\|\cdot\|_\infty)\colon A\mapsto (r_1,\dots,r_m)$, is Lipschitz continuous with Lipschitz constant 2.
 \end{lemma}

 \begin{proof}
 Let $A_1, A_2\in\CA_m$ be standard arcs with vertices $(\bfv_1^1,\dots,\bfv_m^1)=\vertex(A_1)$ and $(\bfv_1^2,\dots,\bfv_m^2)=\vertex(A_2)$ and with scaling factors $(r_1^1,\dots,r_m^1)=\scale(A_1)$ as well as $(r_1^2,\dots,r_m^2)=\scale(A_2)$ respectively. Since for all $i=1,\dots,m$ we have
 \[r_i^1 \leq r_i^2 + \dist\kl{\bfv^1_{i-1}, \bfv^2_{i-1}} + \dist\kl{\bfv^1_{i}, \bfv^2_{i}}
 \]
 and vice versa
 \[r_i^2 \leq r_i^1 + \dist\kl{\bfv^1_{i-1}, \bfv^2_{i-1}} + \dist\kl{\bfv^1_{i}, \bfv^2_{i}},
\]
 we obtain
 \[
    \nkl{\scale\kl{A_1} - \scale\kl{A_2}}_{\infty} = \max_{{i=1,\dots,m}} \bkl{r_i^1 - r_i^2} \leq 2d_\CA\kl{A_1, A_2}. \qedhere
 \]
 \end{proof}

 Altogether we can conclude this section by combining all pieces and proving the local Lipschitz continuity of $\Xi\colon\Omega_{\delta,m}\to\R^m_+$.
 \begin{proof}[Proof of \cref{lem:xi_loclip}]
  The claim follows directly from \cref{asm:main,lem:loclip1,lem:loclip2} since the composition of locally Lipschitz continuous functions is again locally Lipschitz continuous.
 \end{proof}

\section{Proof of the Reverse Result (\cref{thm:counter})}\label{sec:counterproof}
 We will now come to the proof of \cref{thm:counter} and show that exploiting any of the three restrictions in \cref{thm:main} indeed allows us to find ReLU-invariant families of distributions. We split the proof in three parts and construct examples for each of the three cases.

 \subsection{Families of Distributions in One Dimension}\label{sec:counterproof_i}
 In this section we will prove part \ref{itm:counter_i} of \cref{thm:counter}. We explicitly construct a family of ReLU-invariant probability distributions on $\R$ with a locally Lipschitz continuous parametrisation map.

 The key observation is that ReLU neural networks in one dimension are a rather restricted class of functions unlike networks in higher dimensions. In fact, we will show that increasing the depth of neural networks beyond three layers does not lead to the ability to express more complicated functions. Every one-dimensional ReLU neural network can be rewritten as a three layer network. Thus we only require a constant number of parameters to describe the set of all one-dimensional ReLU networks. It is then straightforward to obtain a ReLU-invariant family of probability distributions by explicitly making these parameters part of the family's parametrisation.

 We will follow a similar idea to obtain ReLU-invariant families in higher dimension when proving part \ref{itm:counter_iii} of \cref{thm:counter} in \cref{sec:counterproof_iii}. However in this case the number of parameters to describe the neural networks grows with their depth leading to the need for an arbitrary large number of parameters to describe the distributions in the family. Although it is possible to generalise our construction to higher dimensions this comes at the cost of losing the local Lipschitz continuity of the parametrisation map.

 We proceed in three steps. We first analyse the set of one-dimensional ReLU networks and show that it can be completely described by only six parameters. We then use these to parametrise all ReLU neural networks as functions in $C(K,\R)$ on some compact domain $K\subseteq\R$ via a so called \emph{realisation} map $R\colon\R^6\to C(K,\R)$. Finally, we use a pushforward map $Q\colon C(K,\R)\to\CD(\R)$ mapping neural network functions $f\in C(K,\R)$ to the pushforward of a fixed \emph{prototype} measure $\mu_0\in\CD(\R)$ under $f$ to generate the family of probability distribution. The prototype measure is assumed to be supported within $K$. The one-parameter family $p\colon\R^6\to\CD(\R)$ is then simply given as $p=Q\,\circ R$. We will now discuss each of the steps in more detail.

\begin{figure}
     \centering
     \resizebox{\textwidth}{!}{
     % func(x, a_1, a_2, c_1, c_2)
\pgfmathdeclarefunction{boundedfunc}{5}{%
\pgfmathparse{%
(#1<=#2) * (#4) +%
(#1>=#3) * (#5) +%
and(#1>#2, #1<#3) * (#4 + (#5-#4)/(#3-#2)*(#1-#2))%
}%
}%
% func(x, a, c, w)
\pgfmathdeclarefunction{unboundedfunc}{4}{%
\pgfmathparse{%
(#4*(#1-#2)<0) * (#3) +%
(#4*(#1-#2)>=0) * (#3 + #4*(#1-#2))%
}%
}%
\begin{tabular}{cccc}
  \multicolumn{2}{c}{\textbf{unbounded case}} & \multicolumn{2}{c} {\textbf{bounded case}} \\
  slope $w>0$ & slope $w<0$ &  slope $\frac{c_2-c_1}{a_2-a_1}>0$ & slope $\frac{c_2-c_1}{a_2-a_1}<0$ \\[.5em]
  \begin{tikzpicture}[scale=0.5]
    \begin{axis}[
        axis x line=middle,
        axis y line=middle,
        ymin=-.5, ymax=4, ytick={1}, yticklabel={\LARGE $c$},
        y tick label style={anchor=west, xshift=.5em, black},
        tick style={thin, gray},
        xmin=-5, xmax=5, xtick={-1}, xticklabel={\LARGE $a$},
        x tick label style={anchor=north, yshift=0em, black},
    ]
    \addplot[color0,thick,samples at={-5,-1,5}]{unboundedfunc(x,-1,1,0.5)};
    %\node[rectangle, draw, fill=gray!15, anchor=west] (f) at (-4.5,4) {$f(x)=\begin{cases}c,&\quad w(x-a)\leq 0\\c+w(x-a),&\quad w(x-a)>0\end{cases}$};
    \draw[thin, gray, dashed] (-1,1) -- (-1,0);
    \draw[thin, gray, dashed] (-1,1) -- (0,1);
    \end{axis}
  \end{tikzpicture}
&
  \begin{tikzpicture}[scale=0.5]
    \begin{axis}[
        axis x line=middle,
        axis y line=middle,
        ymin=-.5, ymax=4, ytick={1}, yticklabel={\LARGE $c$},
        y tick label style={anchor=east, xshift=0em, black},
        tick style={thin, gray},
        xmin=-5, xmax=5, xtick={1}, xticklabel={\LARGE $a$},
        x tick label style={anchor=north, yshift=0em, black},
    ]
    \addplot[color0,thick,samples at={-5,1,5}]{unboundedfunc(x,1,1,-0.5)};
    %\node[rectangle, draw, fill=gray!15, anchor=east] (f) at (4.5,4) {$f(x)=\begin{cases}c,&\quad w(x-a)\leq 0\\c+w(x-a),&\quad w(x-a)>0\end{cases}$};
    \draw[thin, gray, dashed] (1,1) -- (1,0);
    \draw[thin, gray, dashed] (1,1) -- (0,1);
    \end{axis}
  \end{tikzpicture}
&
  \begin{tikzpicture}[scale=0.5]
    \begin{axis}[
        axis x line=middle,
        axis y line=middle,
        ymin=-.5, ymax=4, ytick=\empty,
        xmin=-5, xmax=5,  xtick=\empty,
    ]
    \addplot[color0,thick,samples at={-5,-2,1,5}]{boundedfunc(x,-2,1,1,2)};
    % \node[rectangle, draw, fill=gray!15] (f) at (0,3) {$f(x)=\begin{cases}
    %     c_1 &\quad x \leq a_1 \\
    %     c_2,&\quad x \geq a_2 \\
    %     c_1 + \frac{c_2-c_1}{a_2-a_1}(x-a_1),&\quad a_1<x<a_2
    %     \end{cases}$};
    \draw[thin, gray, dashed] (-2,1) -- (-2,0) node[below, black] {\LARGE $a_1$};
    \draw[thin, gray, dashed] (1,2) -- (1,0) node[below, black] {\LARGE $a_2$};
    \draw[thin, gray, dashed] (-2,1) -- (0,1) node[right, black] {\LARGE $c_1$};
    \draw[thin, gray, dashed] (1,2) -- (0,2) node[left, black] {\LARGE $c_2$};
    \end{axis}
  \end{tikzpicture}
&
  \begin{tikzpicture}[scale=0.5]
    \begin{axis}[
        axis x line=middle,
        axis y line=middle,
        ymin=-.5, ymax=4, ytick=\empty,
        xmin=-5, xmax=5,  xtick=\empty,
    ]
    \addplot[color0,thick,samples at={-5,-1,2,5}]{boundedfunc(x,-1,2,2,1)};
    % \node[rectangle, draw, fill=gray!15] (f) at (0,3) {$f(x)=\begin{cases}
    %     c_1 &\quad x \leq a_1 \\
    %     c_2,&\quad x \geq a_2 \\
    %     c_1 + \frac{c_2-c_1}{a_2-a_1}(x-a_1),&\quad a_1<x<a_2
    %     \end{cases}$};
    \draw[thin, gray, dashed] (-1,2) -- (-1,0) node[below, black] {\LARGE $a_1$};
    \draw[thin, gray, dashed] (2,1) -- (2,0) node[below, black] {\LARGE $a_2$};
    \draw[thin, gray, dashed] (-1,2) -- (0,2) node[right, black] {\LARGE $c_1$};
    \draw[thin, gray, dashed] (2,1) -- (0,1) node[left, black] {\LARGE $c_2$};
    \end{axis}
  \end{tikzpicture}
\\
\multicolumn{2}{c}{
\begin{tikzpicture}
    \node[rectangle, draw, fill=gray!15, anchor=west, minimum height=2cm] (f) at (-4.5,4) { \small $f(x)=\begin{cases}c,&\quad w(x-a)\leq 0\\c+w(x-a),&\quad w(x-a)>0\end{cases}$};
  \end{tikzpicture}
}
&
\multicolumn{2}{c}{

\begin{tikzpicture}
    \node[rectangle, draw, fill=gray!15, minimum height=2cm] (f) at (0,3) { \small
    $f(x)=\begin{cases}
        c_1 &\quad x \leq a_1 \\
        c_2,&\quad x \geq a_2 \\
        c_1 + \frac{c_2-c_1}{a_2-a_1}(x-a_1),&\quad a_1<x<a_2
        \end{cases}$};
  \end{tikzpicture}
}
\end{tabular}
     }
     \caption{ReLU networks in dimension $d=1$ can only take one of five distinct forms. We show the four non-constant cases. This includes two unbounded cases (top) and two bounded cases (bottom).}
     \label{fig:onedim_relu}
\end{figure}
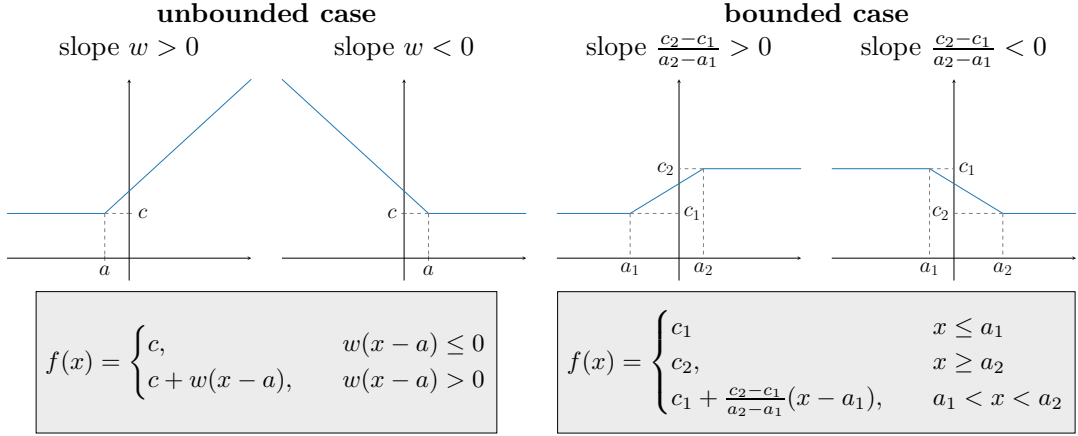

 \paragraph{One-dimensional ReLU networks:} Let us first show that any one dimensional ReLU network can be rewritten as a three layer network. For this let $L\in\N$ and $f\colon\R\to\R$ be an $L$-layer ReLU network given as
 \[
    f=f_L\circ f_{L-1}\circ\dots\circ f_2\circ f_1
  \]
  with layers
  \[
    f_i\colon\R\to\R\colon x\mapsto\varrho(w_i x + b_i)\quad\text{for}\quad i=1,\dots,L.
  \]
  We observe that $f$ is constant as soon as any weight $w_i$ is zero and a constant function can easily be rewritten as a three layer network. So from now on we assume $w_i\neq 0$ for all layers. We denote by
  \[
    A_i = \skl{\,x\in\R\,:\,w_i x + b_i \geq 0\,}\quad\text{for}\quad i=1,\dots,L
  \]
  the domain on which the $i$-th layer is affine linear with slope $w_i$. It is constant on the complement $A_i^c$. Further we denote
  \[
    B_i = \kl{f_{i-1}\circ\dots\circ f_1}^{-1}(A_i)\quad\text{for}\quad i=2,\dots,L
  \]
  and $B_1=A_1$. Each $A_i$ is a convex and closed subset of $\R$ and $f_{i-1}\circ\dots\circ f_1$ is continuous and monotone as a composition of continuous and monotone functions. Hence, also all $B_i$ are closed and convex. In $\R$ these are exactly the (possibly unbounded) closed intervals.
  \begin{lemma}\label{lem:onedim_relu_const}
  Let $U\subseteq \R\setminus\bigcap_{i=1}^L B_i$ be connected. Then $f\vert_U$ is constant.
  \end{lemma}
  \begin{proof}
   We have $\R\setminus \bigcap_{i=1}^L B_i = \bigcup_{i=1}^L B_i^c$ and therefore $U = \bigcup_{i=1}^L U\cap B_i^c$. Since $B_i$ is closed $B_i^c$ is open and thus $U\cap B_i^c$ is relatively open in $U$. But by the choice of $B_i$ we know $\kl{f_i\circ\dots\circ f_1}\kl{U\cap B_i^c} \equiv 0$, hence $f\vert_{U\cap B_i^c}$ is constant. But since $f$ is continuous and $U$ is connected this already implies that $f\vert_U$ is constant.
  \end{proof}
  The one-dimensional ReLU network $f$ is piecewise constant except for the (possibly empty) region $\bigcap_{i=1}^L B_i$ where all layers are non constant and act affine linearly. On this region $f$ is affine linear with slope $\prod_{i=1}^L w_i$. As discussed above each $B_i$ is convex and closed, hence the same holds for their intersection. Two different cases can occur, which determine the overall shape of the function $f$. This is also visualized in \cref{fig:onedim_relu}.

  Firstly, if $\bigcap_{i=1}^L B_i$ is bounded it must be empty, a single point, or a bounded and closed interval. Then $f$ is bounded. In fact it is constant or a piecewise linear step function with one step and two constant regions. These functions can be represented by three layer networks.

  Secondly, if $\bigcap_{i=1}^L B_i$ is an unbounded closed interval, then $f$ looks like a shifted and scaled ReLU function and can again be represented by a three layer ReLU network.

  \clearpage

  \begin{corollary}\label{cor:onedim_relu}
  The ReLU network $f$ has one of the following forms.
  \begin{itemize}
      \item $f$ is bounded and has two constant and one affine linear region:
      \[
        f(x) = \begin{cases}
        c_1, & \quad x \leq a_1 \\
        c_2, & \quad x \geq a_2 \\
        c_1 + \frac{c_2-c_1}{a_2-a_1}(x-a_1), & \quad a_1<x<a_2
        \end{cases}
      \]
      for some $c_1, c_2\geq 0$ and $a_1<a_2$. This includes the degenerate case where $f$ is constant by choosing $c_1=c_2$.
      \item $f$ is unbounded and has one constant and one affine linear region:
      \[
        f(x) = \begin{cases}
        c, & \quad w(x-a) \leq 0 \\
        c + w(x-a), & \quad w(x-a)>0
        \end{cases}
      \]
      for some $c\geq 0$, $a\in\R$, and $w\neq 0$.
  \end{itemize}
  \end{corollary}
  \begin{proof}
   This directly follows from \cref{lem:onedim_relu_const}, the continuity and monotonicity of the ReLU network $f$ and the fact that $f$ is lower bounded by zero.
  \end{proof}
  \begin{lemma}
  The two types of piecewise linear functions from \cref{cor:onedim_relu} can both be written as one dimensional ReLU networks with three layers.
  \end{lemma}
  \begin{proof}
   We give explicit formulas for choosing the weights $w_1, w_2, w_3$ and biases $b_1, b_2, b_3$ of the three layer network.

   In the first case of a bounded function we choose
   \begin{alignat*}{2}
   w_1 &= 1, \qquad&\qquad b_1 &= -a_1, \\
   w_2 &= -\frac{|c_2-c_1|}{a_2-a_1}, \qquad&\qquad b_2 &= |c_2-c_1|,  \\
   w_3 &= -\sign\kl{c_2-c_1}, \qquad&\qquad b_3 &= c_2,
   \end{alignat*}
   and in the second case of an unbounded function we choose
   \begin{alignat*}{2}
   w_1 &= \sign\kl{w}, \qquad&\qquad b_1 &= -\sign\kl{w}a, \\
   w_2 &= |w|, \qquad&\qquad b_2 &= c, \\
   w_3 &= 1, \qquad&\qquad b_3 &= 0.
   \end{alignat*}
   The claim now follows by straightforward calculations.
  \end{proof}

 \paragraph{Neural Network Parametrisation:}
 Neural networks can be viewed algebraically (as collections of weights and biases) or analytically (as functions in some function space). This distinction was made in \cite{pv2018network-calculus} and used in \cite{prv2018nn-topology-arxiv} to study the dependence of neural network functions on their weights and biases. The mapping from weights and biases to the corresponding neural network function is referred to as the \emph{realisation} map. In \cite{prv2018nn-topology-arxiv} the continuity of the realisation map was shown for neural networks of fixed size with continuous activation function (in particular this includes the ReLU activation), when they are viewed as functions in $C(K,\R^d)$ on a compact domain $K$. If the activation function is Lipschitz continuous (which again is the case for the ReLU activation) also the realisation map is Lipschitz continuous.

 As we have seen all ReLU neural networks in one dimension can be rewritten into a three layer network and can thus be parametrised by three weight and three bias parameters. Let $K\subseteq\R$ be any compact non-empty domain. For any collection of weights and biases $\theta=(w_1,b_1,w_2,b_2,w_3,b_3)\in\R^6$ we denote the function that is the ReLU neural network realisation of these weights and biases as $f_\theta\colon K\to\R$. Then the realisation map is
 \[
    R\colon\R^6\to C(K,\R^d)\colon \theta\mapsto f_\theta.
 \]
 From \cite[Proposition 5.1]{prv2018nn-topology-arxiv} we obtain the following result.
 \begin{lemma}
    The realisation map $R\colon\R^6\to C(K,\R^d)$ is Lipschitz continuous.
 \end{lemma}

 \paragraph{Pushforward Mapping:} We want to connect neural network functions with the effect they have on probability measures and obtain an invariant family of distributions. For this let $\mu_0\in\CD(\R)$ be any probability measure on $\R$ with $\supp(\mu_0)\subseteq K$. We call this the \emph{prototype} measure and will derive all other measures in the family as pushforwards of $\mu_0$ under ReLU neural networks. We define the pushforward function
 \[
    Q\colon C(K,\R)\to\CD(\R)\colon f\mapsto f_\ast\mu_0.
 \]
 The restriction of $Q$ to the subset of ReLU neural network functions will be used to generate the invariant family of distributions.
 \begin{lemma}
    The pushforward function $Q\colon C(K,\R)\to\CD(\R)$ is Lipschitz continuous.
 \end{lemma}
 \begin{proof}
   Let $f,g\in C(K,\R)$ and $A\in\CB(\R)$. Assuming $f^{-1}(A)\neq\emptyset$, for any $x\in f^{-1}(A)$ we have $f(x)\in A$ and
   $|f(x)-g(x)|\leq\|f-g\|_\infty$. Consequently, for any $\epsilon>\|f-g\|_\infty$ we get $g(x)\in A^\epsilon$ and therefore $x\in g^{-1}\kl{A^\epsilon}$. Thus $f^{-1}(A)\subseteq g^{-1}\kl{A^\epsilon}$. Conversely, if $f^{-1}(A)=\emptyset$ then the same inclusion trivially holds. Analogously $g^{-1}(A)\subseteq f^{-1}\kl{A^\epsilon}$ can be shown for any $\epsilon>\|f-g\|_\infty$. This directly implies
   \begin{align*}
       d_P(Q(f), Q(g)) &= d_P(f_\ast\mu_0, g_\ast\mu_0)\\
       &= \inf\left\{\,\epsilon >0\,:\, f_\ast\mu_0(A) \leq g_\ast\mu_0(A^\epsilon) + \epsilon \text{ and } \right. \\
       &\qquad\qquad\left. g_\ast\mu_0(A) \leq f_\ast\mu_0(A^\epsilon) + \epsilon \text{ for any }A\in\CB(\R)\,\right\} \\
       &= \inf\left\{\,\epsilon >0\,:\, \mu_0(f^{-1}(A)) \leq \mu_0(g^{-1}(A^\epsilon)) + \epsilon \text{ and } \right. \\
       &\qquad\qquad\left. \mu_0(g^{-1}(A)) \leq \mu_0(f^{-1}(A^\epsilon)) + \epsilon \text{ for any }A\in\CB(\R)\,\right\} \\
       &\leq \|f-g\|_\infty.\qedhere
   \end{align*}
 \end{proof}

 \paragraph{ReLU-Invariance:}
 We can now define the family of distributions as
 \[
    p\colon\R^6\to\CD(\R)\colon \theta \mapsto \kl{Q\circ R}\kl{\theta}.
 \]
 The local Lipschitz continuity of $p$ is clear from the previous steps. It remains to establish the ReLU-invariance.
 \begin{lemma}
 The family of distributions $p$ is ReLU-invariant.
 \end{lemma}
 \begin{proof}
   Let $\theta\in\R^6$ be arbitrary and $p(\theta)\in\CD(\R)$ the corresponding probability measure. Further, let $f\colon\R\to\R\colon x\mapsto\varrho\kl{wx+b}$ for some $w,b\in\R$ be any ReLU-layer. We need to show that there exists a $\omega\in\R^6$ such that $p(\omega)=f_\ast p(\theta)$.

   By construction $p(\theta) = \kl{Q\circ R}(\theta) = \kl{R(\theta)}_\ast\mu_0 = \kl{f_\theta}_\ast\mu_0$, where $f_\theta=R(\theta)$ is a ReLU neural network. Clearly also $f\circ f_\theta$ is a ReLU neural network, and since any ReLU network can be rewritten as a three-layer network, there exists $\omega\in\R^6$ such that $R(\omega)=f\circ f_\theta$. Finally,
   \[
        p(\omega)=(f\circ f_\theta)_\ast \mu_0 = f_\ast \kl{f_\theta}_\ast \mu_0 = f_\ast p(\theta)
    \]
    yields the claim.
 \end{proof}

 \subsection{Families of Distributions with Finite Support}\label{sec:counterproof_ii}
 In this section we will prove part \ref{itm:counter_ii} of \cref{thm:counter} and show that ReLU-invariant families of probability distributions exist in which all distributions are finitely supported.
 A finitely supported Radon probability measure $\mu\in\CD(\R^d)$ can be expressed as a mixture of finitely many Dirac measures,
 \[
 \mu = \sum_{i=1}^{N} c_i \delta_{\bfa_i},
 \]
 for some $N\in\N$, $\bfa_i\in\R^d$, and $0\leq c_i \leq 1$ with $\sum_{i=0}^N c_i = 1$. For any measurable function $f\colon\R^d\to\R^d$ the pushforward of such a mixture is simply the mixture of the individual pushforwards, that is
 \[
    f_\ast\kl{\mu} =  \sum_{i=1}^{N} c_i f_\ast\kl{\delta_{\bfa_i}}.
 \]
 The pushforward of a Dirac measure is again a Dirac measure, more precisely we simply have $f_\ast\kl{\delta_{\bfa_i}} = \delta_{f(\bfa_i)}$, and therefore
 \[
    f_\ast\kl{\mu} =  \sum_{i=1}^{N} c_i \delta_{f(\bfa_i)}.
 \]
 In particular, for a ReLU-layer $f\colon\R^d\to\R^d\colon\bfx\mapsto\varrho\kl{\bfW\bfx+\bfb}$ we get
  \[
    f_\ast\kl{\mu} =  \sum_{i=1}^{N} c_i \delta_{\varrho\kl{\bfW\bfa_i+\bfb}}.
 \]
 Altogether, this shows that for any $N\in\N$ the set
 \[
    \CD_N = \skl{\,\mu\in\CD(\R^d)\,:\,|\supp(\mu)|\leq N\,}
 \]
 is ReLU-invariant in any dimension $d\in\N$. It can be described by $n=N(d+1)$ parameters by simply specifying all the locations $\bfa_i$ and mixture coefficients $c_i$. More precisely, let $\Delta_N=\skl{\,\bfc\in[0,1]^N\,:\,\sum_i c_i=1\,}$ and
 \[
 \Omega_N=\underbrace{\R^d\times\dots\times\R^d}_{N\text{ times}}\times\Delta_N\subseteq \R^n
 \]
 and
 \[
 p_N\colon\Omega_N\to\CD(\R^d)\colon\kl{\bfa_1,\dots,\bfa_N,\bfc}\mapsto \sum_{i=1}^{N} c_i \delta_{\bfa_i}.
 \]
 Then $p_N$ is Lipschitz continuous and parametrises $\CD_N$.

 \subsection{Families of Distributions Without Local Lipschitz Continuity}\label{sec:counterproof_iii}
 In this section we will prove part \ref{itm:counter_iii} of \cref{thm:counter}. If we omit the local Lipschitz continuity of the parametrisation we can construct a ReLU-invariant one-parameter family of distributions in $\R^d$ for any dimension $d\in\N$. We proceed similar to the ideas in \cref{sec:counterproof_i}, however we need to take care of the fact that the set of all ReLU networks can not easily be described by a constant number of parameters, unlike in the one-dimensional case.

 We define a family of distributions each of which can be described by a finite but arbitrarily large number of parameters. We use space-filling curves to fuse these arbitrarily many parameters into a single parameter.

 The main idea is quite straightforward and analogous to \cref{sec:counterproof_i}: If we have an arbitrarily large number of parameters available to describe each distribution in the family we can simply use these parameters to specify the weights of the neural network transformations with respect to which we want to obtain invariance. The challenge in the general $d$-dimensional setting is to show that this leads to a continuous parametrisation.

 We start by clarifying what we mean by a finite but arbitrarily large number of parameters. We use the conventional notation $\R^\omega$ for the product of countably many copies of $\R$ equipped with the product topology. In other words, $\R^\omega$ is the space of all real-valued sequences. Further we denote by
 \[
 \R^\infty = \skl{\,\kl{x_1,x_2,\dots}\in\R^\omega\,:\,x_i\neq 0\text{ for only finitely many }i\,}
 \]
 the subset of sequences that are eventually zero. The space $\R^\infty$ will serve as the parameter space. Each element in $\R^\infty$ effectively uses only a finite number of parameters (since all the remaining ones are zero), however this number can be arbitrarily large.

 We proceed in three steps. We first introduce a one-parameter description $\Gamma\colon\R\to\R^\infty$ of the finitely-but-arbitrarily-many parameter set $\R^\infty$. We then use a subset $\Omega_\infty\subseteq\R^\infty$ to parametrise ReLU neural networks as functions in $C(K,\R^d)$ using a realisation map $R\colon\Omega_\infty\to C(K,\R^d)$. Finally, we use a pushforward map $Q\colon C(K,\R^d)\to\CD(\R^d)$ mapping neural network functions $f\in C(K,\R^d)$ to the pushforward of a fixed \emph{prototype} measure $\mu_0\in\CD(\R^d)$ under $f$ to generate the family of probability distribution. As before the prototype measure is assumed to be supported within the domain $K$. The one-parameter family $p\colon\Omega\to\CD(\R^d)$ is then given as $p=Q\,\circ R\,\circ\Gamma$ for some suitable domain $\Omega$. We will now discuss each of the steps in more detail.

 \paragraph{Space-Filling Curves:}
 We want to continuously describe $\R^\infty$ with a single parameter. This can be achieved by gluing together continuous space-filling curves from the unit interval to cubes of arbitrary dimension. The existence of such maps is guaranteed by the Hahn-Mazurkiewicz theorem, see for example \cite[Chapter 6.8]{Sagan1994spacefilling}.
 \begin{lemma}\label{lem:space-filling}
 There exists a continuous and surjective function $\Gamma\colon\R\to\R^\infty$.
 \end{lemma}
 The construction of $\Gamma$ is deferred to \ref{apx:space-filling}.

 \paragraph{Neural Network Parametrisation:}
 We need to extend the realisation map from \cref{sec:counterproof_i} to arbitrary dimensions and arbitrary numbers of layers. Further we will show that the results in \cite{prv2018nn-topology-arxiv} concerning continuous network parametrisations can be extended to networks of arbitrary depth.

 A ReLU neural network in $d$ dimensions is characterised by the number of its layers as well as $d^2+d$ parameters for the weights and biases for each of the layers. Using the first component in $\R^\infty$ to encode the number of layers we can parametrise all such neural networks with the set
 \[
    \Omega_\infty = \skl{\,\kl{L,x_1,x_2,\dots}\in\R^\infty\,:\,L\in\N\text{ and }x_i=0\text{ for all }i>L\kl{d^2+d}\,} \subseteq \R^\infty.
 \]
 It will be convenient to introduce notations for the subsets of parameters with a fixed number of layers. For $L\in\N$ we write $\Omega_L = \skl{L}\times \R^{L\kl{d^2+d}}\times\skl{0}^\infty\subseteq\Omega_\infty$ and observe that
 \[
    \Omega_\infty = \bigcup_{L\in\N} \Omega_L.
 \]
 In fact this is the partition of $\Omega_\infty$ into its connected components.

 For any $\theta=(L,x_1,x_2,\dots)\in\Omega_\infty$ we can regroup its leading non-zero components into alternating blocks of size $d^2$ and $d$ and write \[\theta=(L,\bfW_1,\bfb_1,\bfW_2,\bfb_2,\dots,\bfW_L,\bfb_L,0,\dots)\]
 with $\bfW_i\in\R^{d\times d}$ and $\bfb_i\in\R^d$. As before let $K\subseteq\R^d$ be any non-empty compact domain and denote the function that is the ReLU neural network realisation of the weights $\skl{\bfW_1,\dots,\bfW_d}$ and biases $\skl{\bfb_1,\dots,\bfb_d}$ as $f_\theta\colon K\to\R^d$. Then we can define the extended realisation map
 \[
    R\colon\Omega_\infty\to C(K,\R^d)\colon \theta\mapsto f_\theta.
 \]
 \begin{lemma}
    The extended realisation map $R\colon\Omega_\infty\to C(K,\R^d)$ is continuous.
 \end{lemma}
 \begin{proof}
  To show the continuity of $R\colon\Omega_\infty\to C(K,\R^d)$ it suffices to show the continuity on all connected components $\Omega_L$ of the domain. But for each fixed $L\in\N$ we know from \cite[Proposition 5.1]{prv2018nn-topology-arxiv} that the realisation map is continuous from $\Omega_L$ to $C(K,\R^d)$. This already proves the overall continuity of $R$.
 \end{proof}

 \paragraph{Pushforward Mapping:} We need to extend our pushforward map from \cref{sec:counterproof_i} to arbitrary dimension. As before let $\mu_0\in\CD(\R^d)$ be any probability measure on $\R^d$ with $\supp(\mu_0)\subseteq K$. We define the pushforward function analogous to before as
 \[
    Q\colon C(K,\R^d)\to\CD(\R^d)\colon f\mapsto f_\ast\mu_0.
 \]
 The restriction of $Q$ to the subset of ReLU neural network functions will be used to generate the invariant family of distributions.
 \begin{lemma}
    The pushforward function $Q\colon C(K,\R^d)\to\CD(\R^d)$ is continuous.
 \end{lemma}
 The proof works exactly as in the one-dimensional case in \cref{sec:counterproof_i}.

 \paragraph{ReLU-Invariance:}
 We can now define the one-parameter family of distributions as
 \[
    p\colon\Omega\to\CD(\R^d)\colon \theta \mapsto \kl{Q\circ R\circ\Gamma}\kl{\theta}
 \]
 with the domain $\Omega=\Gamma^{-1}(\Omega_\infty)$ chosen so that $\Gamma$ maps it exactly to the feasible neural network parameters $\Omega_\infty$ and thus $R\circ\Gamma$ maps it to all ReLU neural networks in $d$ dimensions. The continuity of $p$ is clear from the three previous steps. It remains to establish the ReLU-invariance.
 \begin{lemma}
 The one-parameter family of distributions $p$ is ReLU-invariant.
 \end{lemma}
 \begin{proof}
   Let $\theta\in\Omega$ be arbitrary and $p(\theta)\in\CD(\R^d)$ the corresponding probability measure. Further, let $f\colon\R^d\to\R^d\colon\bfx\mapsto\varrho\kl{\bfW\bfx+\bfb}$ for some $\bfW\in\R^{d\times d}$ and $\bfb\in\R^d$ be any ReLU-layer. We need to show that there exists a $\omega\in\Omega$ such that $p(\omega)=f_\ast p(\theta)$.

   By construction we have that $p(\theta) = \kl{Q\circ R\circ \Gamma}(\theta) = \kl{R(\Gamma(\theta))}_\ast\mu_0 = \widetilde{f}_\ast\mu_0$, where $\widetilde{f}=R(\Gamma(\theta))$ is a ReLU neural network. Clearly also $f\circ \widetilde{f}$ is a ReLU neural network, so there exists $\omega\in\Omega$ such that $R(\Gamma(\omega))=f\circ \widetilde{f}$. Finally,
   \[
        p(\omega)=(f\circ \widetilde{f})_\ast \mu_0 = f_\ast \widetilde{f}_\ast \mu_0 = f_\ast p(\theta)
    \]
    yields the claim.
 \end{proof}

\section{Discussion}\label{sec:discussion}
Our work gives a comprehensive characterisation of distribution families that are invariant under transformations by layers of ReLU neural networks. The only invariant distributions are either sampling distributions or rather degenerate and elaborately constructed distributions that are infeasible for practical applications. This justifies the use of approximation schemes such as Assumed Density Filtering.

We have limited our analysis to the class of functions defined as layers of ReLU neural networks. Similarly, one could ask the same question for other function classes.

Considering different commonly used activation functions, such as $\tanh$, the logistic function, or the Heaviside function, instead of ReLUs is one variation that comes to mind. Since the Heaviside function is not continuous and has a discrete and finite range, it transforms any distribution into a distribution with finite support. So clearly also in this case the only invariant distributions can be sampling distributions. For smooth activation functions, such as $\tanh$ and the logistic function, we are not aware of any characterisation of invariant distributions. Extending our proof strategy to this scenario is not straightforward as it relies on specific properties of the ReLU. We leave this question open for future research.

A second variation on the class of functions would be to put restrictions on the weight matrices or bias vectors. This could be either general constraints, for example non-negativity or positive-definiteness, or more specific restrictions by allowing the weight matrices only to be chosen from a small set of allowed matrices. Both kinds of constraints arise in the
context of iterative reconstruction methods for solving inverse and sparse coding problems \cite{hoyer2002nonneg} or in corresponding unrolled and learnable iterative algorithms \cite{gregor_learning_2010,pock2017varnets}. Here, the restrictions on the weight matrices often stem from physical constraints of the model describing the inverse problem.
We comment on both aspects of this variation in \ref{apx:restricted_weights}.
A brief investigation shows that for any finite or even countable set of weights and biases there exist indeed parametrisations of distributions that circumvent the three restrictions (R1)--(R3) in \cref{thm:main}. These, however, are purely of theoretical interest, since they rely on calculating infinitely many distributions a priori and are thus not useful in practice.

\section*{Acknowledgements}
We thank Peter Jung for providing helpful feedback during the process of writing the first draft of this manuscript, as well as Mones Raslan and John Sullivan for fruitful discussions during the earlier stages of preparing this work.
J. M. and S. W. acknowledge partial support by DFG-GRK-2260 (BIOQIC).

\clearpage

%%%%% References %%%%%
\bibliographystyle{abbrv}
\bibliography{references,ref-arcs}

%%%%% APPENDIX %%%%%
\appendix

\section{Metric Spaces and Hausdorff Dimension}\label{apx:hdim}
In this section we will proof \cref{lem:loclip_n_to_m}. For this we will first review some concepts and results regarding general metric spaces and their Hausdorff dimension, which we denote by $\dim_{H}$. Applying these to Euclidean $\R^n$ or subspaces thereof will yield the desired result.

A topological space is called a Lindel{\"o}f space if every open cover of it has a countable subcover. This is weaker than compactness, where the existence of finite subcovers is required. For metric spaces the notions Lindel{\"o}f, separable, and second-countable are all equivalent. It is easy to see that any $\sigma$-compact space is Lindel{\"o}f. Subspaces of separable metric spaces are again separable. We start with a collection of useful properties of the Hausdorff dimension, see for example \cite[Chapter 6]{Edgar2008fractalgeometry}.

\begin{lemma}\label{lem:hdim_props}
Let $(X,d_X)$ and $(Y,d_Y)$ be metric spaces, $A,B\subseteq X$ Borel sets, $\kl{A_i}_{i\in\N}$ a countable collection of Borel sets $A_i\subseteq X$, and $f\colon X\to Y$ (globally) Lipschitz continuous. Then
\begin{enumerate}[label=(\roman*)]
  \item $A\subseteq B \Rightarrow \dim_{H}(A)\leq \dim_{H}(B)$,
  \item $\dim_{H}(A\cup B) = \max\skl{\dim_{H}(A), \dim_{H}(B)}$,
  \item $\dim_{H}\kl{\bigcup_{i\in\N} A_i} = \sup_{i\in\N}\skl{\dim_{H}(A_i)}$,
  \item $\dim_{H}(f(X))\leq \dim_{H}(X)$.
\end{enumerate}
\end{lemma}

If $(X, d_X)$ is a separable space, then the last part can be generalised to locally Lipschitz continuous maps.

\begin{lemma}\label{lem:hdim_loclip}
Let $(X,d_X)$ and $(Y,d_Y)$ be metric spaces, $f\colon X\to Y$ locally Lipschitz continuous, and $X$ separable. Then
\[\dim_{H}(f(X))\leq \dim_{H}(X).\]
\end{lemma}
\begin{proof}
For every $x\in X$ there exists an open neighbourhood $U_x$ of $x$ such that $f$ restricted to $U_x$ is Lipschitz continuous. Also $(U_x)_{x\in X}$ is an open cover of $X$. As $X$ is separable and thus Lindel{\"o}f there exists a countable subcover $(U_{x_i})_{i\in \N}$. From \cref{lem:hdim_props} we conclude $\dim_{H}(f(U_{x_i}))\leq\dim_{H}(U_{x_i})\leq\dim_{H}(X)$ for all $i\in\N$. Since $f(X) = f(\bigcup_{i\in\N} U_{x_i}) = \bigcup_{i\in\N} f(U_{x_i})$, we can again use \cref{lem:hdim_props} to conclude
\[\dim_{H}(f(X)) = \dim_{H}(\bigcup_{i\in\N}f(U_{x_i})) \leq \sup_{i\in\N} \dim_{H}(f(U_{x_i})) \leq \dim_{H}(X).\qedhere\]
\end{proof}

We now come to the special case of the Euclidean space $\R^n$.
Every Borel subset of $\R^n$ with non-empty interior has Hausdorff dimension $n$, again see for example \cite[Chapter 6]{Edgar2008fractalgeometry}. A direct consequence of this is the following result.
\begin{lemma}\label{lem:hdim_Rn}
We have $\dim_{H}(\R^n) = n$ and $\dim_{H}(\R^n_+)=n$.
\end{lemma}

Let us now restate and proof \cref{lem:loclip_n_to_m}.
\hdimlemma*
\begin{proof}
 Since $B\subseteq \R^n$ is a separable metric space, we can use \cref{lem:hdim_loclip,lem:hdim_Rn} to obtain
  \[m=\dim_H(f(B))\leq \dim_H(B)\leq \dim_H(\R^n) = n.\qedhere\]
\end{proof}

\section{Space-Filling Curves in Arbitrary Dimensions}\label{apx:space-filling}
 In this section we describe the construction of the continuous and surjective function $\Gamma\colon\R\to\R^\infty$ and thus prove \cref{lem:space-filling}.

 We start with constructing a map from the unit interval to the $n$-dimensional cube for any $n\in\N$. The existence of such maps is guaranteed by the Hahn-Mazurkiewicz theorem, see for example \cite[Chapter 6.8]{Sagan1994spacefilling}. We can then glue these maps together to obtain $\Gamma$.

 Recall that $\R^\omega$ is the product of countably many copies of $\R$ equipped with the product topology. It can be thought of as the space of all real-valued sequence. Further, $\R^\infty$ denotes the subset of all real-valued sequences that are eventually zero.

 Let $g\colon\ekl{0,1}\to \ekl{0,1}^2$ be any continuous and surjective space-filling curve. Examples for such curves are the Sierpi\'nski curve, the Hilbert curve, or the Peano curve, see \cite{Sagan1994spacefilling} for an overview of various space-filling curves. We extend this to higher-dimensional cubes by iteratively defining curves $g_n\colon\ekl{0,1}\to\ekl{0,1}^n$ for $n\geq 2$ as
 \begin{align*}
     g_2 &= g \\
     g_n &= \kl{\text{id}_{n-2}\times g} \circ g_{n-1},\qquad n\geq 3,
 \end{align*}
 which are again continuous and surjective. From these we can obtain continuous and surjective curves $h_n\colon\ekl{0,1}\to\ekl{-n,n}^n$ from the unit interval to the scaled symmetric $n$-dimensional cubes by scaling and translating. Since we ultimately want glue all these $h_n$ together we also want to assure $h_n(0)=h_n(1)=\bfzero$ for all $n\geq 2$. Thus we define
 \[
    h_n(t) =\begin{cases}
                4nt\kl{2g_n(0)-\bfone}, &\quad t\in\ekl{0,\frac{1}{4}}, \\[.5em]
                n\kl{2g_n\kl{2t-\frac{1}{2}}-\bfone}, &\quad t\in\ekl{\frac{1}{4},\frac{3}{4}},  \\[.5em]
                4n(1-t)\kl{2g_n(1)-\bfone}, &\quad t\in\ekl{\frac{3}{4},1}.
            \end{cases}
 \]
 Finally, we glue the pieces together to get a continuous map from one real parameter to any arbitrary finite number of parameters. We define the map $\Gamma\colon \R \to \R^\omega$ that maps the interval $\ekl{n, n+1}$ surjectively to the $\ekl{-n,n}^n$ cube on the first $n$ coordinates of $\R^\omega$, that is
 \[
  \Gamma(t) =
  \begin{cases}
  h_{\floor{t}}(t-\floor{t})\times \skl{0}^{\infty}, &\quad t\geq 2, \\
  \skl{0}^{\infty} &,\quad t< 2.
  \end{cases}
 \]
 The only critical points regarding the continuity of $\Gamma$ are the integers where we transition from one interval to the next and thus $\floor{t}$ changes. But by assuring $h_n(0)=h_n(1)=\bfzero$ for all $n\geq 2$ we achieve a continuous gluing at the interval transitions. Thus $\Gamma$ is continuous restricted to each of the intervals $(-\infty,2], [2, 3], [3, 4], \dots$ respectively. These form a locally finite cover of $\R$ by closed sets, hence we can use the pasting Lemma, see for example \cite[Chapter III.9]{Dugundji1966pasting}, to conclude the continuity of $\Gamma$ on all of $\R$. The map is not surjective onto $\R^\omega$. However for our purpose we only require surjectivity onto $\R^\infty$ and this is clearly satisfied, since
 \[
  \R^\infty = \bigcup_{n\geq 2} [-n,n]^n\times\skl{0}^\infty = \bigcup_{n\geq 2} \Gamma([n, n+1]).
 \]

 \section{Restricting the Set of Weight Matrices}\label{apx:restricted_weights}
 In this section we briefly discuss some variations of characterisations of $\CF$-invariant families of distributions, where $\CF$ is not the entire set of all ReLU layers.
 If the considered collection $\CF$ of functions is much more restricted, it might be possible to obtain invariant families of distributions.

\subsection{General Restrictions of the Weight Matrices}
The setting we considered so far has no restrictions on the weight matrices. We make use of that in our constructive proof by repeatedly relying on rotation and projection matrices.
Putting restrictions on the matrices, such as non-negativity, symmetry, positive-definiteness, or allowing only diagonal matrices, would prohibit our proof strategy. Using only diagonal matrices renders the functions in $\CF$ separable in their input components and thus effectively reduces to the one-dimensional case. For this we have already discussed the invariant distributions in \cref{sec:counterproof_i}. For all other matrix restrictions it remains to be investigated whether our approach can be adapted.

\subsection{Restrictions on the Number of Matrices and Biases}

Another practical concern can be the number of different possible weight matrices. So far we considered a continuum of matrices and biases. One possible restriction is to consider a finite collections $\CF$ instead.

We begin with the extreme case, in which the collection contains only a single measurable function $\CF=\{f\colon\R^d\to\R^d\}$, for example a ReLU layer $f(\bfx) = \varrho\kl{\bfW\bfx + \bfb}$ with a fixed weight matrix $\bfW$ and bias vector $\bfb$. As in \cref{sec:counterproof_i,sec:counterproof_iii} we can construct an invariant family of distributions starting with a prototype distribution $\mu_0\in\CD(\R^d)$. Next, we iteratively define
\begin{equation*}
    \mu_n = f_\ast \mu_{n-1},\quad n\in\N,
\end{equation*}
and using $\eta(t) = t-\floor{t}$ also the intermediate interpolations
\begin{equation}\label{eq:meas_interpol}
    \mu_{t} = (1-\eta(t))\mu_{\floor{t}} + \eta(t)\mu_{\floor{t}+1},\quad t\geq 0.
\end{equation}
We quickly show the Lipschitz continuity of $t\mapsto\mu_t$.
Let $0\leq t_1 < t_2 < \infty$. Without loss of generality, we can assume $t_2 \leq t_1 + 1$, since the Prokhorov metric of two probability distributions is always bounded by $1$.
First, we consider the case $\floor{t_2} = \floor{t_1} = m$. Then
\begin{align*}
    d_P(\mu_{t_2},\mu_{t_1})
    &\leq \nkl{\mu_{t_2}-\mu_{t_1}}_{TV} \\
    & \leq \nkl{ (\eta(t_2)-\eta(t_1))(\mu_{m+1}-\mu_{m})}_{TV} \\
    & \leq\bkl{ \eta(t_2)-\eta(t_1)}\nkl{\mu_{m+1}-\mu_{m}}_{TV} \\
    &\leq \bkl{t_2-t_1}\kl{\nkl{\mu_{m+1}}_{TV}+\nkl{\mu_{m}}_{TV}} \\
    & =2 \bkl{t_2-t_1},
\end{align*}
where $\nkl{\cdot}_{TV}$ is the total variation norm.

Second, we consider the case $\floor{t_2} = \floor{t_1} + 1 = m$. Then
\begin{align*}
    \mu_{t_2}-\mu_{t_1} = (1-\eta(t_2)-\eta(t_1))\mu_{m} + \eta(t_2)\mu_{m+1} - (1-\eta(t_1))\mu_{m-1}
\end{align*}
and since
\[
    \bkl{1-\eta(t_2)-\eta(t_1)} \leq \bkl{1-\eta(t_1)}+\bkl{\eta(t_2)} = t_2 - t_1
\]
we obtain
\begin{align*}
   d_P(\mu_{t_2},\mu_{t_1}) &\leq \nkl{\mu_{t_2}-\mu_{t_1}}_{TV} \\
   &\leq \bkl{1-\eta(t_2)-\eta(t_1)} + \bkl{\eta(t_2)} + \bkl{1-\eta(t_1)}\\
   &\leq 2 \bkl{t_2-t_1}.
\end{align*}
Thus, the one-parameter mapping $t\mapsto\mu_t$ is Lipschitz continuous. It also satisfies $f_\ast \mu_{t} = \mu_{t+1}$ for any $t\geq 0$, which means that it is $f$-invariant (hence $\CF$-invariant).

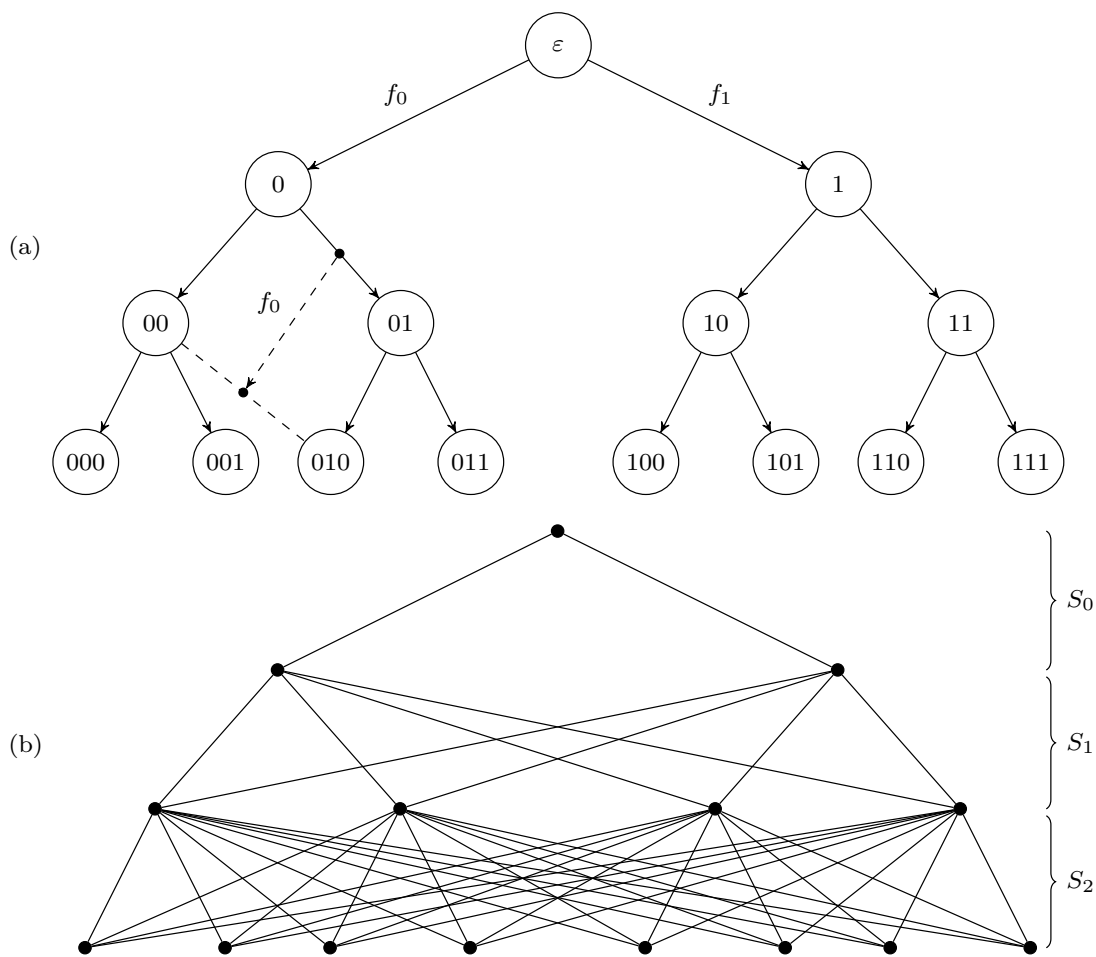
\begin{figure}
    \centering
    \resizebox{\textwidth}{!}{
    \scriptsize
\begin{tabular}{@{}l@{\,\,}l@{}}
(a) & 
\raisebox{-.5\height}{\begin{tikzpicture}[scale=0.8,->,>=stealth',
level 1/.style={sibling distance = 8cm,
  level distance = 2cm},
level 2/.style={sibling distance = 3.5cm,
  level distance = 2cm},
  level 3/.style={sibling distance = 2cm,
  level distance = 2cm}] 
\node [mstyle] (mu) {$\varepsilon$}
    child{ node [mstyle] (v0) {$0$} 
            child{ node [mstyle] (v00) {$00$} 
            	child{ node [mstyle] (bottom)  {$000$}}
            	%edge from parent node[above left] {$x$}} %for a named pointer
							child{ node [mstyle]{$001$}}
            }
            child{ node [mstyle] (v01) {$01$}
							child{ node [mstyle] (v010) {$010$}}
							child{ node [mstyle] {$011$}}
            }
            edge from parent 
            node[anchor=south east] {$f_0$}
        }
    child{ node [mstyle] {$1$}
            child{ node [mstyle] {$10$} 
							child{ node [mstyle] {$100$}}
							child{ node [mstyle] {$101$}}
            }
            child{ node [mstyle] {$11$}
							child{ node [mstyle] {$110$}}
							child{ node [mstyle] {$111$}}
            }
        edge from parent 
        node[anchor=south west] {$f_1$}
		}
	;
	
	\node[draw, fill, circle, inner sep=1pt] (vt) at ($(v0)!0.5!(v01)$) {};
	\draw[dashed, thin,-] (v00) -- (v010);
	\node[draw, fill, circle, inner sep=1pt] (vtf) at ($(v00)!0.5!(v010)$) {};
	\draw[dashed, thin] (vt) -- (vtf) node[midway,above left] {$f_0$};
\end{tikzpicture}} 
\\ \\
(b) &
\raisebox{-.5\height}{\begin{tikzpicture}[scale=0.8,-,>=stealth',
level 1/.style={sibling distance = 8.0cm,
  level distance = 2cm},
level 2/.style={sibling distance = 3.5cm,
  level distance = 2cm},
  level 3/.style={sibling distance = 2.0cm,
  level distance = 2cm}]
\node [vstyle] (eps) {}
    child{ node [vstyle] (v0) {} 
            child{ node [vstyle] (v00) {} 
            	child{ node [vstyle] (v000) {}}
							child{ node [vstyle] (v001) {}}
            }
            child{ node [vstyle] (v01) {}
							child{ node [vstyle] (v010) {}}
							child{ node [vstyle] (v011) {}}
            }                            
    }
    child{ node [vstyle] (v1) {}
            child{ node [vstyle] (v10) {}
							child{ node (v100) [vstyle] {}}
							child{ node (v101) [vstyle] {}}
            }
            child{ node [vstyle] (v11) {}
							child{ node (v110) [vstyle] {}}
							child{ node (v111) [vstyle] {}}
            }
		}
;
\node[opacity=0.0, minimum width=0.75, circle] (phantom) at (v000) {000};
\draw[] (v0)--(v10); \draw[] (v0)--(v11);
\draw[] (v1)--(v00); \draw[] (v1)--(v01);

\draw[] (v00)--(v100); \draw[] (v00)--(v101); \draw[] (v00)--(v110);
\draw[] (v00)--(v111); \draw[] (v00)--(v011); \draw[] (v00)--(v010);

\draw[] (v01)--(v100); \draw[] (v01)--(v101); \draw[] (v01)--(v110);
\draw[] (v01)--(v111); \draw[] (v01)--(v001); \draw[] (v01)--(v000);

\draw[] (v10)--(v000); \draw[] (v10)--(v001); \draw[] (v10)--(v110);
\draw[] (v10)--(v111); \draw[] (v10)--(v011); \draw[] (v10)--(v010);

\draw[] (v11)--(v000); \draw[] (v11)--(v001); \draw[] (v11)--(v100);
\draw[] (v11)--(v101); \draw[] (v11)--(v011); \draw[] (v11)--(v010);

\draw [decorate,decoration={brace,amplitude=3pt,raise=3pt},xshift=11cm]
(v111.east|-eps) -- (v111.east|-v1) node [black,midway,xshift=0.5cm] {$S_0$};

\draw [decorate,decoration={brace,amplitude=3pt,raise=3pt},xshift=11cm]
(v111.east|-v1)++(0,-0.1) -- (v111.east|-v11) node [black,midway,xshift=0.5cm] {$S_1$};

\draw [decorate,decoration={brace,amplitude=3pt,raise=3pt},yshift=0pt]
(v111.east|-v11)++(0,-0.1) -- (v111.east) node [black,midway,xshift=0.5cm] {$S_2$};
\end{tikzpicture}}
\end{tabular}
    }
    \caption{(a) Consecutively applying either $f_0$ or $f_1$ to a prototype measure $\mu$ results in an infinite tree structure (only the first three levels are shown). Transforming an intermediate (interpolated) point by $f_0$ or $f_1$ (dashed) results in a point outside the tree structure. (b) Adding additional edges to represent also the transformations of all interpolated points results in a multipartite graph in which all vertices of consecutive levels are connected. The subgraphs $S_i$ induced by two consecutive levels of vertices are complete bipartite graphs.}
    \label{fig:treegraph}
 \end{figure}

This idea can also be extended to finite (or even countable) collections $\CF$. For brevity we only illustrate the concept for a collection $\CF=\{f_0, f_1\}$ of two functions. Again, we start by choosing an arbitrary prototype measure, which we will simply denote $\mu$ in this case.
For any binary string $\bfz\in \skl{0,1}^\ast$, where $\varepsilon$ refers to the empty string, we define
\begin{align*}
  \mu_{\varepsilon} &= \mu, \\
  \mu_{\bfz} &= \kl{f_{z_n}\circ\dots\circ f_{z_1}}_\ast \mu.
\end{align*}
The procedure of obtaining measures this way can be associated to a perfect binary tree of infinite depth with root $\varepsilon$. A binary string $\bfz_1$ is the child of another string $\bfz_2$ if it extends it by exactly one digit $0$ or $1$, see \cref{fig:treegraph}. Note that different strings can result in the same measure so that the set of measures $\mu_{\bfz}$ is not in one-to-one correspondence to the vertices of the tree. However, each measure is represented by at least on vertex.

In order to find a parametrisation that includes all the measures $\mu_{\bfz}$ and that is locally Lipschitz continuous and $\CF$-invariant, we will turn the tree into an undirected graph. The parametrisation will extend the interpolation idea from \eqref{eq:meas_interpol} and results from a walk through all nodes of the graph.

Let $l(\bfz)$ denote the length of a binary string $\bfz$. We define the graph $G=(V,E)$ with vertices $V=\skl{0,1}^\ast$ and edges $E = \skl{ (\bfz_1, \bfz_2) \in V\times V\,:\,\bkl{l(\bfz_1)-l(\bfz_2)} = 1}$, i.e. every vertex is a binary string and all strings of consecutive lengths are connected by an edge. The resulting graph is a multipartite graph where the independent sets are strings of the same length.
Let us define subgraphs $S_i$ of $G$ for $i\in \N_0$ that are induced by vertex sets
\[
  V_i = \skl{0,1}^i \cup \skl{0,1}^{i+1}.
\]
The resulting subgraphs are complete bipartite graphs, see \cref{fig:treegraph}.
We define a walk $\CW \in \skl{V}^\ast$ on $G$ as a sequence of vertices where two consecutive vertices are connected by an edge.
We are now looking for an infinite walk on $G$ that passes through all edges at least once.

Since we can pass an edge multiple times such a walk is easy to construct: Each subgraph $S_i$ has only finitely many edges, which means there exists a walk that passes through them (except for $S_0$ we can even find an Eulerian cycle for each subgraph). Without loss of generality we can assume that the walk on $S_i$ starts and ends in $\bfzero_i$ (the string of $i$ zeros) and denote it $\CW_i$. Let $\sqcap$ symbolise the concatenation of two walks. We set
\[
 \CW = \bigsqcap_{k=0}^{\infty} \CW_i,
\]
which is possible since in $G$ the two vertices $\bfzero_i$ and $\bfzero_{i+1}$ are connected by an edge for every $i\in\N$. Denoting the $i$-th vertext visited by $\CW$ as $\CW(i)$ we can now define
\begin{equation}\label{eq:walk_parametrisation}
  \mu_t = (1-\eta(t))\mu_{\CW(\floor{t})} + \eta(t)\mu_{\CW(\floor{t}+1)},
\end{equation}
with $\eta(t)$ as before.

To see that this fulfils our criteria consider the following. All distributions in the parametrised family are of the form
\[
  (1-s) \mu_{\bfz_1} + s \mu_{\bfz_2}, \quad \text{with}\quad \bfz_1 \in \skl{0,1}^k, \bfz_2 \in \skl{0,1}^{k+1}, k\in\N_0, s\in[0,1].
\]
Transforming it by the function $f_i$ for $i\in\skl{0,1}$ results in the pushforward measure
\[
  (1-s) \mu_{\bfz^{\prime}_1} + s \mu_{\bfz^{\prime}_2}, \quad\text{with}\quad \bfz^{\prime}_1 = (\bfz_1, i), \bfz^{\prime}_2 = (\bfz_2, i).
\]
But since $\bfz^{\prime}_1$ and $\bfz^{\prime}_2$ still have length difference $1$, they share an edge in $S^{k+1}$.
The walk $\CW$ passes through all edges, so we can define $m\in \N$ as the smallest number where either $\CW_m = \bfz^{\prime}_1$ and $\CW_{m+1} = \bfz^{\prime}_2$ or $\CW_m = \bfz^{\prime}_2$ and $\CW_{m+1} = \bfz^{\prime}_1$. Let us without loss of generality assume that the former holds. Then
\[
 (1-s) \mu_{\bfz^{\prime}_1} + s \mu_{\bfz^{\prime}_2} = \mu_{m+s},
\]
which shows $t\mapsto\mu_t$ is $\CF$-invariant. The Lipschitz continuity follows analogously to the case with only a single  function above.

This idea can even be extended to countably many different measurable functions.
Instead of a single binary tree and corresponding graph $G$, we could define a sequence of trees $T_i$ and corresponding graphs $G_i$ where $T_i$ is $i$-ary, corresponding to the first $i$ functions.
Every tree $T_i$ is a subtree of $T_j$ whenever $j>i$ and correspondingly the associated graph $G_i$ is a subgraph of $G_j$. Thus any walk on $G_i$ is also valid on $G_j$.
If now $\CW_i$ denotes a walk on $G_i$ starting at the root $\varepsilon$, covering every edge up to the $i$-th level in $G_i$, and returning back to the root $\varepsilon$, then we can set
\[
 \CW = \bigsqcap_{k=1}^{\infty} \CW_i.
\]
It follows that every edge of every graph in the sequence will eventually be reached. Defining the parametrisation as in \cref{eq:walk_parametrisation}, ensures that both $\CF$-invariance as well as Lipschitz continuity holds.

 \end{document}